\documentclass[preprint,12pt]{elsarticle}

\usepackage{lineno}

\usepackage{xspace}
\usepackage{pgfgantt}
\usepackage{amsmath}
\usepackage{amssymb}
\usepackage{enumitem}
\setlist[enumerate]{itemsep=0mm}
\usepackage{booktabs}
\usepackage{multirow}
\usepackage{tabularx}
\usepackage{hyperref}
\usepackage{pdfpages}
\usepackage{graphicx,subcaption}

\newcommand{\bm}[1]{\mathbf{#1}}

\hypersetup{
    colorlinks=true,
    linkcolor=blue, 
}

\journal{Reliability Engineering $\&$ System Safety}

\begin{document}

\begin{frontmatter}

\title{Disentangling Slow and Fast Temporal Dynamics in Degradation Inference with Hierarchical Differential Models}
%
%
%

\author{Mengjie Zhao}
\ead{mengjie.zhao@epfl.ch}

\author{Olga Fink}
\ead{olga.fink@epfl.ch}

\address{Swiss Federal Institute of Technology Lausanne (EPFL), Switzerland}






\newcommand{\revtext}[1]{{\color{black}#1}}


\begin{abstract}
Reliable inference of system degradation from sensor data is fundamental to condition monitoring and prognostics in \revtext{mechanical and infrastructural} systems. Since degradation is rarely \revtext{directly} observable and measurable, it must be inferred \revtext{without any access to the true degradation state} to enable accurate health assessment and decision-making.
This is particularly challenging because operational and environmental variations dominate system behavior, while degradation introduces only subtle, long-term changes. Consequently, sensor data primarily reflect short-term operational variability, making it difficult to disentangle the underlying degradation process.
\revtext{Most unsupervised degradation inference methods learn nominal system behavior and use the residual as proxies for degradation. However, these} residuals remain strongly entangled with operational history, \revtext{leading to} noisy and unreliable degradation estimation, particularly in \revtext{infrastructural} systems with \revtext{with dominant transient dynamics}. 
Neural Ordinary Equations (NODEs) offer a \revtext{flexible} framework for \revtext{modeling} latent dynamics, but \revtext{in degraded systems they suffer from} numerical stiffness and degradation disentanglement remains difficult.
To address these \revtext{challenges}, we propose a novel Hierarchical Controlled Differential Equation (H-CDE) framework \revtext{that jointly models slow degradation dynamics and fast operational dynamics within a unified architecture. H-CDE improves numerical efficiency through separate time integration of slow and fast components. Through the proposed} learnable path transformation that maps raw inputs to a latent degradation\revtext{-relevant control path together with} a novel \revtext{monotonicity enforcing} activation function that explicitly regularizes the inferred degradation dynamics, H-CDE enables effective disentangled degradation inference. 
Comprehensive evaluations on both \revtext{mechanical and infrastructural} systems \revtext{demonstrate} that H-CDE outperforms residual-based baselines, yielding more accurate, robust, and interpretable \revtext{degradation} inference \revtext{in a fully unsupervised setting}.

\end{abstract}


\begin{highlights}
\item \revtext{Unsupervised} latent states align well with true physical degradation.
\item \revtext{H-CDE} shows robust generalization to unseen conditions.
\item The primary latent component serves as an interpretable health indicator.
\item \revtext{Effective} degradation \revtext{disentanglement both in mechanical and structural systems}.
\end{highlights}





\end{frontmatter}

%

\newcommand{\revtext}[1]{{\color{black}#1}}

\section{Introduction}
%
%
%
%

Ensuring the reliability and safety of complex engineered systems, ranging from critical infrastructure~\cite{zaparolicunha_review_2023} to industrial machinery~\cite{pimenov_artificial_2023} and aerospace structures~\cite{schwartz_unsupervised_2022}, \revtext{requires both condition monitoring at the individual-system level and reliability analysis at the fleet level. 
Condition monitoring focuses on tracking the health evolution of a system using sensor measurements, while reliability analysis aims to quantify fleet-level failure risk, lifetime distributions, and maintenance strategies, often without direct access to the true system health state.}
A key \revtext{link between these two levels is} degradation, \revtext{which provides a continuous measure of health evolution and supports} early fault detection and \revtext{maintenance decision-making}. 
\revtext{Monitoring degradation evolution enables the identification of abnormal acceleration associated with fault onset. When multiple run-to-failure trajectories are available, degradation can also support downstream reliability tasks such as} Remaining Useful Life (RUL) prediction~\cite{ahmedmurtaza_paradigm_2024, fink_potential_2020}. 

\revtext{While certain systems offer measurable physical proxies for health estimation, such as capacity in battery systems~\cite{liu_capacity_2024}, obtaining accurate measurements typically requires controlled operating conditions. In contrast, degradation is not directly measurable in most mechanical and structural systems. Although physical relationships exist between degradation and measurable quantities (e.g., stiffness loss and displacement in bridges, or efficiency loss and throttle angle in turbofan engines), these relationships are strongly influenced by operational and environmental variability. Consequently, degradation remains a latent process and must be estimated or inferred from monitored sensor data during operation. 
Existing approaches differ primarily in their data requirements, in particular, whether degradation labels or full run-to-failure trajectories are required for training, as well as in whether they target fleet-level reliability analysis or individual asset-level health assessment. In this work, methods that rely on explicit degradation information are referred to as \textit{degradation estimation} approaches, whereas methods that infer degradation as a latent variable directly from sensor data without such labels are referred to as \textit{degradation inference}. These approaches can be broadly classified into three families: reliability-based, physics-based, and data-driven methods~\cite{zagorowska_survey_2020, jarosz_recent_2025}.}



\textit{Reliability-based} methods model the macroscopic progression of degradation as stochastic processes\revtext{, leveraging fleet-level data from multiple units.} Common choices include Gamma processes, Wiener processes\revtext{~\cite{liu_capacity_2024, he_physicsinformed_2025}}, and Weibull-type accelerated life models~\cite{shahraki_review_2017}. \revtext{While effective for fleet-level reliability analysis, they are less suited for estimating degradation for individual assets, especially when degradation is not directly observable and difficult to measure~\cite{zhang_degradation_2023}.}
\revtext{Both physics-based and data-driven} methods aim to infer degradation by modeling how it affects system dynamics under different operating conditions. \revtext{\textit{Physics-based} methods apply when the governing dynamics are known or can be reasonably approximated.}
These models incorporate prior knowledge of system behavior, and the hidden degradation state can be inferred from sensor measurements using filtering techniques such as Kalman filters~\cite{tian_realtime_2022} or particle filters~\cite{liu_degradation_2024b}.
In many real-world systems, however, the governing physics is unknown or too complex to model accurately. In such cases, \textit{data-driven} methods are increasingly \revtext{adopted} for their ability to \revtext{jointly learn} system dynamics and \revtext{infer} latent degradation states directly from \revtext{condition monitoring} data~\cite{balali_datadriven_2020}.

\revtext{Most \textit{data-driven} methods model degradation in a supervised manner, typically through RUL prediction tasks~\cite{pepe_neural_2022, zhou_timevarying_2023}, where degradation evolution is captured implicitly rather than estimated explicitly.}
\revtext{When explicit degradation labels are unavailable, }residual-based modeling is \revtext{a common alternative}. 
\revtext{These methods} learn a representation of \revtext{healthy} system behavior and interpret deviations from this reference as indicators of degradation~\cite{zhou_systematic_2024, hsu_comparison_2023}. \revtext{Although primarily developed for condition monitoring and anomaly detection, the magnitude of the residuals is often used as a proxy for health indicators~\cite{sarwar_probabilistic_2024}.}
However, the inferred degradation is often noisy. \revtext{Residual-based} methods assume that \revtext{sensor measurements} from \revtext{a} degraded system can be \revtext{decomposed} into a nominal \revtext{healthy} response and \revtext{an} additive degradation\revtext{-induced} deviation. In practice, the impact of degradation on system behavior is highly nonlinear~\cite{si_nonlinear_2022}. \revtext{As a result, residuals capture not only degradation effects but also} unmodeled nonlinear \revtext{interactions with the operating history}, making it difficult to isolate the degradation \revtext{from operational variability}.


In addition, degradation manifests differently across engineering domains. 
For high-cycle components or subsystems such as batteries~\cite{navidi_physicsinformed_2024}, aero-engines~\cite{fu_prognostic_2023}, or gearboxes~\cite{su_advanced_2024}, system 
\revtext{dynamics are often assumed to evolve on a much faster time scale than degradation. 
As a result, the system can be approximated as operating close} to equilibrium at each operating point. 
In these \textit{steady-state systems}, degradation primarily affects efficiency \revtext{or performance} and can be modeled as an additional \revtext{latent} input influencing the mapping from exogenous inputs to sensor measurements~\cite{ariaschao_combining_2021}. Most data-driven degradation inference methods are \revtext{developed under this assumption}.
In contrast, infrastructure systems such as bridges or wind turbines \revtext{are subject to} dynamic loading and \revtext{and exhibit} continuous response. In these systems, degradation  \revtext{is mainly driven} by material deterioration such as stiffness loss due to cracks or corrosion~\cite{jarosz_recent_2025}. 
\revtext{Although degradation evolves slowly, the sensor measurements correspond to persistent transient responses rather than steady operating points.}
We refer to such systems as \textit{dynamic-response systems}. 
\revtext{For this class of systems, only limited work exists on degradation inference from condition monitoring data, with most studies focusing instead on damage detection.}
\revtext{In this setting,} residual-based methods struggle to isolate degradation due to strong nonlinear entanglement between degradation and \revtext{transient} system dynamics~\cite{sarwar_probabilistic_2024}. To \revtext{better capture dynamic system behavior}, recent methods attempt to model degradation-aware operational dynamics by explicitly conditioning system behavior on the inferred degradation state~\cite{biggio_ageingaware_2023, garmaev_deep_2024}. \revtext{However, these methods primarily target steady-state systems, where dynamics is driven by changes in operating or environmental conditions rather than by intrinsic transient responses, which is substantially more challenging to learn.}

\revtext{Without} direct supervision or reliable physical priors, inferred latent variables \revtext{may capture} operational or environmental variability \revtext{rather than the true degradation}~\cite{locatello_challenging_2019}, \revtext{motivating the need for }a strong inductive bias \revtext{for disentanglement}. In \revtext{practice}, the underlying physical model \revtext{governing operation and degradation} is \revtext{typically} unknown, and available prior knowledge is \revtext{limited to the fact that underlying} degradation\revtext{, such as material fatigue in structural components and internal wear in mechanical components is irreversible for most mechanical and structural systems and evolves according to} the operational \revtext{usage}.
To incorporate this structure, we propose a method that jointly models both the operational dynamics conditioned on degradation and the degradation progression driven by system usage. 
We propose to infer the degradation state through a progression model constrained to be monotonically increasing and use it as a latent variable to guide the learning of operational dynamics. By modeling these two processes simultaneously, the degradation dynamics act as a regularizer and provides a strong inductive bias for learning complex, transient system behavior. This coupling enables the model to capture \revtext{transient} dynamic responses while isolating a physically meaningful latent degradation representation.

To achieve this, we first need to infer the underlying system dynamics from condition monitoring data. Neural Ordinary Differential Equations (NODEs) offer a promising framework by parameterizing continuous-time dynamics using neural networks~\cite{kidger_neural_2022}. This yields smooth latent trajectories, which are well-suited for capturing long-range temporal dependencies. However, standard NODEs are limited in their ability to model systems influenced by \revtext{exogenous inputs such as} control inputs or external factors. Since their dynamics are defined solely by the initial condition, the system's evolution during inference remains fixed and cannot adapt to changes in external inputs over time. 
Neural Controlled Differential Equations (NCDEs)~\cite{kidger_neural_2020} overcome this limitation by treating the input data stream as a control signal that drives the system. 
This enables NCDEs to incorporate the full operational history and evolve dynamically in response to changing \revtext{exogenous inputs}.

However, degrading systems introduce additional complexity because they exhibit multiscale dynamics, with degradation progressing much more slowly than operational dynamics. While NCDEs effectively model controlled continuous-time systems, they are primarily designed for single-timescale behavior. When applied to multiscale systems, standard NODE-based models (including NCDEs) suffer from numerical stiffness due to the large disparity in time scales between fast and slow processes. Mitigating this stiffness requires extremely small integration steps, which significantly increases computational cost and may even prevent the model from learning altogether~\cite{meijer_numerical_2011, shi_efficient_2025}.

To address these challenges, we propose the Hierarchical Controlled Differential Equation (H-CDE) framework for disentangled degradation inference. H-CDE disentangles slow degradation dynamics from fast operational fluctuations through a two-layer hierarchical structure: a high-level module learns long-term degradation progression, while a lower-level module captures short-term operational dynamics conditioned on the inferred degradation state.
To mitigate numerical stiffness, H-CDE adopts a multi-scale integration scheme, using coarse integration steps for the slow dynamics and finer steps for the fast dynamics. To prevent contamination of the slow degradation process by high-frequency operational noise, a learnable transformation extracts latent degradation drivers aligned with the degradation timescale in the slow layer. Finally, we enforce monotonic degradation progression through a custom activation function in the differential operator, which acts as a regularizer for disentanglement.

We evaluate H-CDE on two different case studies: a bridge structure with \textit{dynamic-response} and a turbofan engine operating under quasi-steady conditions. These case studies demonstrate the framework’s broad applicability in disentangling latent degradation states across systems with fundamentally different dynamical behavior. To the best of our knowledge, this is the first study to compare degradation inference across such fundamentally different dynamical regimes, whereas prior work has primarily focused on steady-state systems.
Our contributions include: 
\begin{itemize}[leftmargin=1.5em, itemsep=4pt, topsep=4pt, partopsep=0pt, parsep=0pt]
\item \revtext{A fully \revtext{unsupervised framework for degradation inference} that does not require degradation labels or run-to-failure trajectories.}
\item \revtext{The first extension of NCDEs to degradation inference, overcoming the limitation of NODE not being able to generalize to new operating conditions after the differential operator is learned.}
\item \revtext{A novel path transformation and activation function for the} differential operator \revtext{that enable} degradation disentanglement.
\item \revtext{To our knowledge, the first degradation inference framework demonstrated to be applicable to both mechanical and infrastructure systems.}
\end{itemize}

The remainder of the paper is organized as follows. Section~\ref{sec:related_work} reviews related work. Section~\ref{sec:preliminaries} introduces the necessary preliminaries. The proposed H-CDE framework is presented in Section~\ref{sec:method}. Section~\ref{sec:case-studies} describes the case studies, followed by the experimental setup in Section~\ref{sec:exp_setup}. Results and discussions are provided in Section~\ref{sec:results}, and conclusions along with future directions are presented in Section~\ref{sec:conclusion}.

\section{Related Works}
\label{sec:related_work}
This section reviews prior works on \revtext{\textit{unsupervised}} degradation inference methods \revtext{in settings where degradation is hard or even impossible to observe directly, or expensive to measure, as is common in mechanical and infrastructure systems. 
Accordingly, we restrict the review to methods that do not require degradation labels or run-to-failure supervision.} \revtext{We then review} NODE-based methods for latent dynamics learning, with a focus on long-range dependencies and multi-scale dynamics, both crucial for degradation inference, \revtext{and explicitly discuss their application to degradation modeling.}

\subsection{\revtext{Unsupervised} Degradation Inference}
Degradation inference from \revtext{condition monitoring} sensor data \revtext{is an} essential task in Prognostics and Health Management (PHM)~\cite{nguyen_feature_2022}. 
\revtext{In PHM, degradation inference, health indicator construction, and RUL prediction are closely related tasks, as they all rely on extracting degradation-related latent information from sensor measurements.}
The central challenge \revtext{lies in} disentangling \revtext{the} latent degradation state from noise and operational variations. 
\revtext{In the following}, we review \revtext{representative unsupervised approaches} addressing degradation inference, \revtext{including methods that introduce latent variables implicitly interpreted as degradation during training.}

\revtext{\textbf{Residual-based modeling}.}
\revtext{A} common \revtext{unsupervised} approach \revtext{infers degradation indirectly through residuals between observed sensor measurements and learned healthy system behavior.}
\revtext{Such dynamics models} can be \revtext{learned} by mapping \revtext{exogenous input} to \revtext{sensor measurement} using Feedforward Neural Networks (FNNs)~\cite{hsu_comparison_2023}, or by \revtext{employing} Autoencoders (AEs) and Variational Autoencoders (VAEs), \revtext{where degradation is reflected by} the reconstruction error~\cite{zhou_construction_2022}. The resulting residuals \revtext{are often} aggregated into a scalar health indicator, for instance, through dimensionality reduction methods~\cite{zhou_systematic_2024}. 
\revtext{To improve the quality and robustness of inferred degradation indicators, several extensions} incorporate domain knowledge or physical principles through constraints, \revtext{including} monotonicity and trendability constraints in AEs~\cite{bajarunas_health_2024}, trend constraints in VAEs~\cite{qin_unsupervised_2022}, and semi-supervised approaches with isotonic projections \revtext{to enforce monotonic degradation behavior}~\cite{frusque_semisupervised_2024}.
\revtext{However, these methods typically operate at discrete time points, with monotonicity enforced only via loss regularization, resulting in relatively weak constraints. While often sufficient for steady-state systems, it becomes challenging to infer degradation for infrastructure systems reliably due to the strong degradation entanglement with history-dependent transient dynamics.
Sarwar et al.~\cite{sarwar_probabilistic_2024} partially address this limitation by reconstructing sequences of sensor measurements using a probabilistic autoencoder for bridge health monitoring. However, this approach does not impose explicit degradation constraints, and the inferred damage indices remain highly sensitive to ambient temperature, which strongly influences transient dynamics and leads to entanglement between degradation and operating dynamics.

\revtext{\textbf{Degradation-aware operating dynamics}.}
Other advanced architectures explicitly model degradation-aware operating dynamics.} For instance, Garmaev et al.~\cite{garmaev_deep_2024} integrate deep Koopman operators \revtext{to model the degradation dynamics, using exogenous} inputs as additional inputs to reconstruct observed sensor states \revtext{for aeroengines and Computer Numerical Control (CNC) milling machines.} \revtext{However, this formulation operates in discrete time and is primarily suited for steady-state systems.}
Biggio et al.~\cite{biggio_ageingaware_2023} use Transformer encoders to embed ageing states from short voltage/current windows and condition transformer decoders to reconstruct full discharge curves under varying loads. 
\revtext{This approach is demonstrated on battery systems, where measurements are low-dimensional and strongly correlated.
As a result, the conditioned sequence-to-sequence mapping may not readily scale to systems with high-dimensional sensor data or pronounced nonlinear transient dynamics.}

\subsection{Neural Ordinary Differential Equations (NODEs)}
NODEs~\cite{chen_neural_2018, rubanova_latent_2019} and their extension, Neural Controlled Differential Equations (NCDEs)~\cite{kidger_neural_2020}, \revtext{model} system evolution \revtext{through explicit} continuous-time \revtext{differential equations. By parameterizing the underlying differential operators, NODEs provide a structured representation of system dynamics that facilitates the incorporation of prior knowledge and constraints directly on the evolution of latent states.
We review how these models have been used to address long-range temporal dependencies and multi-scale dynamics, and how they have been applied to degradation modeling.}

\textbf{Long-range temporal dependencies}.
Early NODE implementations often \revtext{struggled} with long‐range dependencies due to training instability, but recent advances have addressed this limitation. For instance, rough‐path CDEs use log‐signatures as statistics to describe how the signal drives the CDE and improve memory over long horizons \revtext{on patient medical data}~\cite{morrill_neural_2021}, Neural Laplace models dynamics in the Laplace domain for more efficient, long‐term prediction, \revtext{primarily evaluated on synthetic dynamical systems}~\cite{holt_neural_2022}. \revtext{In addition,} integrating accelerated optimization schemes like Nesterov method with adaptive solvers markedly enhances convergence and stability during training~\cite{nguyen_improving_2022}. \revtext{These developments primarily target long-horizon forecasting tasks and are rarely studied in the context of degradation inference or PHM.}

\textbf{Multiscale dynamics modeling}.
\revtext{NODEs parametrize ODEs that must be solved numerically. They} are rarely applied to multiscale systems, as \revtext{time scale separation} often induces stiffness, \revtext{a classical numerical issue arising when state variables evolve on widely separated time scales. In such cases, standard explicit ODE solvers require extremely small step sizes to maintain stability, leading to inefficient or unstable training, while implicit solvers substantially increase computational cost~\cite{hairer_solving_1996}.}
\revtext{Only} limited efforts \revtext{address these challenges}, for example, by developing multi-scale integration schemes \revtext{to improve neural solvers for stiff systems}~\cite{caldana_neural_2024} or employing hierarchical NODEs tailored to distinct predefined timescales (e.g., minute vs. hour) \revtext{for better long-range financial forecasting}~\cite{qiu_multiscale_2025a}. \revtext{However, these} approaches \revtext{mainly focus on learning coupled slow-fast dynamics and long-range forecasting, and do not consider inferring unobserved slow states from fast states}.

\revtext{\textbf{NODE in degradation modeling}.}
\revtext{Existing NODE-based approaches in PHM predominantly target RUL prediction, where degradation is treated as an implicit latent variable optimized to improve end-of-life prediction rather than as a primary inference objective.}
\revtext{For instance}, Pepe et al.~\cite{pepe_neural_2022} combine augmented Neural ODEs with predictor-corrector RNNs to forecast cycle-level state of health and end-of-life for battery systems. 
Similarly, Zhou et al.~\cite{zhou_timevarying_2023} \revtext{formulate RUL prediction as a constrained,} ODE\revtext{-governed trajectory optimization problem, using a super-network with reinforcement learning} driven \revtext{architecture to adaptively map observations to instantaneous degradation rates}. 
\revtext{While effective, these approaches require run-to-failure labelswhich are often unavailable for systems with long operational lifetimes, such as infrastructure systems, which are often unavailable for systems with long operational lifetimes, such as infrastructure systems.
Moreover, standard NODE formulations assume fixed dynamics once learned, limiting their ability to handle varying operating and environmental conditions. NCDEs explicitly address this by modeling input-driven dynamics; however, to the best of our knowledge, they have not yet been explored for RUL prediction or degradation inference.}

\section{Preliminaries}
\label{sec:preliminaries}
Many real-world processes, including the operation and degradation of industrial and infrastructure assets, evolve continuously over time and depend on both internal state and external inputs. To capture such behavior, this section introduces two foundational frameworks for learning continuous-time dynamics: Neural Ordinary Differential Equations (NODEs) and their extension, Neural Controlled Differential Equations (NCDEs).

\subsection{Neural ODEs}
Neural Ordinary Differential Equations (NODEs) \cite{chen_neural_2018} provide a framework for modeling the continuous evolution of a hidden state $\bm z(t) \in \mathbb{R}^d$. The core idea is to parameterize the time derivative of the state using a neural network $f_\theta:\mathbb{R}^d \times \mathbb{R} \rightarrow \mathbb{R}^d$ with parameters $\theta$:
\begin{equation}
\label{eq:node}
\frac{d\bm z(t)}{dt} = f_\theta(\bm z(t), t).
\end{equation}
Given an initial state  $\bm{z}_0$ at time $t_0$, the trajectory $\bm z(t)$ for $t>t_0$ is obtained by solving this ordinary differential equation (ODE). NODEs are particularly well-suited for modeling autonomous systems where the dynamics explicitly depend only on the current state and time. However, this means that once $f_\theta$ is learned, the solution of NODE is determined by the initial condition $z_0$; there is no direct mechanism to incorporate data that arrives after the initial time point.
To address this limitation, Neural Controlled Differential Equations (NCDEs) \cite{kidger_neural_2020} provide a more flexible framework that naturally adjusts the trajectory based on subsequent observations. 

\subsection{Neural CDEs}
\label{ssec:ncde}
Neural Controlled Differential Equations (NCDEs) \cite{kidger_neural_2020} provide a framework for modeling the continuous-time evolution of a hidden state $\bm z(t) \in \mathbb{R}^{d}$  as it responds to a continuous control path \(\bm X(t) \in \mathbb{R}^m \). 
The dynamics are defined by the controlled differential equation:
\begin{equation}
\label{eq:neural_cde_differential}
d\mathbf{z}(t) = f_\theta\bigl(\mathbf{z}(t)\bigr) d\mathbf{X}(t).
\end{equation}
Here, $f_\theta: \mathbb{R}^d \to \mathbb{R}^{d \times m}$ is a neural network, parameterized by $\theta$, mapping the current state 
$\mathbf{z}(t)$ to a matrix. The term $f_\theta\bigl(\mathbf{z}(t)\bigr) d\mathbf{X}(t)$ represents matrix-vector multiplication, determining the infinitesimal change in the state in response to an infinitesimal change in the control path.
Integrating Eq.~\eqref{eq:neural_cde_differential} over a time interval $[t_0,t]$ gives the state's trajectory:
\begin{equation}
\mathbf{z}(t) = \mathbf{z}(t_0) + \int_{t_0}^t f_\theta\bigl(\mathbf{z}(s)\bigr) d\bm {X}(s).
\label{eq:neural_cde_integral}
\end{equation}
This integral is a Riemann-Stieltjes integral, and allows the hidden state $\bm z(t)$ to evolve in response to the trajectory of the driving signal $\bm X(t)$.

Assuming the observed discrete time series $\mathbf{x} = \{(t_0, \bm{x}_0), \dots, (t_N, \bm{x}_N)\}$ (with $\bm{x}_i \in \mathbb{R}^n$) is a discretizations of an underlying continuous process $\bm X(t)$, we can then approximate $\bm X(t)$ by interpolating the observations $\mathbf{x}$, for instance, using natural cubic splines \cite{kidger_neural_2020}. This yields a smooth path $\bm X(t): [t_0, t_N] \to \mathbb{R}^{m}$ satisfying $\bm X(t_i) = (\bm{x}_i, t_i)$, where time is appended as an additional channel ($m=n+1$). The differentiability of this interpolated path allows the CDE (Eq.~\eqref{eq:neural_cde_differential}) to be expressed as an equivalent ODE:
\begin{equation}
\label{eq:ncde_ode_form}
\frac{d\mathbf{z}(t)}{dt} = f_\theta\bigl(\mathbf{z}(t)\bigr) \frac{d\mathbf{X}(t)}{dt},
\end{equation}
which directly expresses the rate of change of the latent state as a function of the rate of change of the driving signal. 

While this formulation is well-suited for modeling dynamics driven by a single control path, standard NCDEs do not inherently account for coupled dynamics across distinct time scales, such as the multi-scale behavior often found in degradation processes.

\section{Problem Formulation and Proposed Framework}
\label{sec:method}
\subsection{System Dynamics and Degradation Inference}
\label{ssec:slow_fast}
\revtext{We adopt} the \revtext{general state-space formulation discussed in the survey by} Zagorowska et al.~\cite{zagorowska_survey_2020}, representing the degraded system as a coupled slow-fast dynamical system. \revtext{While their survey focuses on control synthesis, we adapt the structure for degradation inference by omitting the explicit output mapping $\mathbf{y}(t)$ and aggregating all exogenous factors (e.g., control inputs and environmental conditions) into a generalized input vector $\mathbf{u}(t)$. 
This system is thus governed by} two coupled processes: (1) the progression of degradation driven by historical usage, and (2) the operating dynamics that are degradation-aware:
\begin{equation}  \label{eq:unified_slow_fast}
\begin{aligned}  
\frac{d\mathbf{d}(t)}{dt} &= \epsilon \,g\left(\mathbf{x}(t), \mathbf{u}(t), \mathbf{d}(t)\right),\\
\frac{d\mathbf{x}(t)}{dt} &= f \big(\mathbf{x}(t), \mathbf{u}(t), \mathbf{d}(t)\big),
\end{aligned}
\end{equation}
where $\mathbf{d}(t) \in \mathbb{R}^m$ denotes the slow degradation state, $\mathbf{x}(t)\in \mathbb{R}^n$ the fast operating states. \revtext{The vector} $\mathbf{u}(t)\in \mathbb{R}^p$ \revtext{is defined as general inputs exogenous to the internal system states that capture the time-varying operating profile.
For example, in bridge monitoring, $\mathbf{u}(t)$ includes both controllable excitations (e.g., traffic loads) and environmental conditions (e.g, ambient temperature), and can induce different vibration or displacement responses $\dot{\mathbf{x}}(t)$ depending on stiffness loss encoded in $\mathbf{d}(t)$. Here $\mathbf{d}(t)$ acts as a slowly varying parameter that modulates the fast dynamics in Eq.~\eqref{eq:unified_slow_fast}}. 
The function $g(\cdot)$ models degradation accumulation, while $f(\cdot)$ governs \revtext{operating dynamics}. The small perturbation parameter $0 < \epsilon \ll 1$ enforces a separation of time scales, with degradation evolving slowly at $O(1/\epsilon)$ compared to $O(1)$ for fast operational dynamics. This formulation captures how $\mathbf{d}(t)$ reflects long-term effects under rapidly changing operating conditions. 

Within this framework, the degradation state $\mathbf{d}(t)$ is the primary quantity of interest for health assessment and prognostics, but is typically unobservable. We therefore define the problem of \textbf{Degradation Inference}: estimating $\mathbf{d}(t)$ from discrete measurements of the observable fast states $\mathbf{x}(t)$ and \revtext{exogenous} inputs $\mathbf{u}(t)$.

\subsection{Hierarchical Neural CDE Framework Overview}
\label{ssec:framework_overview}
\begin{figure}[tbh]
  \centering
    \includegraphics[width=1.\linewidth]{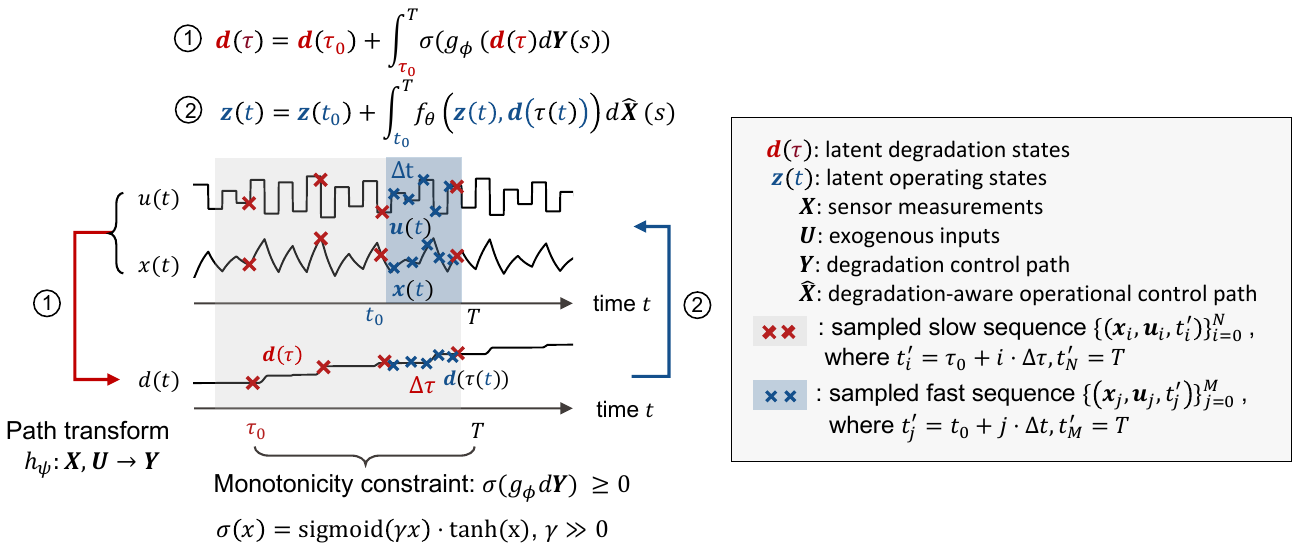}
\caption{\revtext{Architectural} overview of the Hierarchical \revtext{Controlled Differential Equation} (H-CDE) framework \revtext{for disentangled degradation inference from fast operating dynamics}.}


\label{fig:framework}
\end{figure}
The slow-fast dynamic system in Eq.~\eqref{eq:unified_slow_fast} provides the structural basis for modeling degradation, but the underlying functions $f$ and $g$ are unknown and must be learned from data. This is challenging because degraded systems are inherently \textit{multiscale}, making it difficult to infer both latent dynamics and disentangle them.
To address this, we propose the \textbf{Hierarchical Controlled Differential Equation (H-CDE)} framework, a hierarchical model based on NCDEs as introduced in Sec.~\ref{ssec:ncde}. \revtext{An overview of the framework is shown in Fig.~\ref{fig:framework}, which illustrates how the slow–fast dynamics formulated in Eq.~\eqref{eq:unified_slow_fast} are modeled within the H-CDE architecture.}
The H-CDE framework disentangles slow-fast dynamics \revtext{by employing two 
hierarchical layers operating on distinct timescales ($\tau, t$).} 
The slow CDE models \revtext{latent} degradation \revtext{state $\mathbf{d}(\tau)$ over an extended horizon $[\tau_0, T]$ with a} a coarser \revtext{sampling interval $\Delta \tau$. Its evolution is driven by degradation control path $\mathbf{Y}$, derived via the path transformation $h_\psi$, and constrained by a monotonicity regularizer $\sigma(g_{\phi}d\mathbf{Y})\ge0$, which serves as an inductive bias enforcing irreversible degradation progression.} 
\revtext{Subsequently}, the fast layer captures short-term operational dynamics of \revtext{the operational state $\mathbf{z}(t)$ over shorter intervals $[t_0, T]$ with} a finer time resolution, \revtext{driven by a degradation-aware operational control path $\hat{\mathbf{X}}$ that incorporates the previously interpolated degradation state $\mathbf{d}(\tau(t))$ to capture nonlinear wear effects. By enforcing $\tau_0 \ll t_0$ and $\Delta \tau \gg \Delta t$, the H-CDE framework effectively mitigates numerical stiffness through this sequential, decoupled integration.}

The remainder of this section elaborates on the key components \revtext{of the framework}: 
Sec.~\ref{ssec:multi_scale_integration} presents the \textbf{multi-scale time integration} scheme; Sec.~\ref{ssec:control_paths} introduces the \textbf{degradation path transformation} for constructing the slow degradation path $\mathbf{Y}(\tau)$; and Sec.~\ref{ssec:monotonicity} describes the \textbf{monotonicity constraint} enforced for degradation evolution via a novel activation function.

\subsection{Multi-Scale Time Integration}
\label{ssec:multi_scale_integration}
H-CDE aims to learn the underlying dynamics defined in Eqs.~\eqref{eq:unified_slow_fast}. However, learning such latent dynamics is challenging for Neural CDEs. Once the neural vector fields are parameterized, integrating and solving the coupled slow-fast system becomes difficult due to the stiffness from timescale separation~\cite{meijer_numerical_2011}. This stiffness often enforces the use of prohibitively small integration steps or computationally expensive implicit solvers~\cite{shi_efficient_2025}. As a result, learning the neural vector fields end-to-end from data becomes particularly challenging.

To address this, we adopt the quasi-steady-state approximation~\cite{kokotovi?_singular_1999}, which exploits the inherent slowness of the degradation dynamics $\mathbf{d}(t)$ relative to the fast operational dynamics. This allows the slow trajectory to be computed independently and treated as quasi-static when solving for the fast dynamics, yielding the following decoupled integral forms:
\begin{align}
\mathbf{d}(\tau) &= \mathbf{d}(\tau_0) + \int_{\tau_0}^{\tau} g_\phi(\mathbf{d}(\tau)) \, d\mathbf{Y}(s) 
\quad \text{for} \, \tau \in (\tau_0, T] \label{eq:slow_cde_integral}, \\
\mathbf{z}(t) &= \mathbf{z}(t_0) + \int_{t_0}^{t} f_\theta(\mathbf{z}(t), \mathbf{d}(\tau(t))) \,d\hat{\mathbf{X}}(s) 
\quad \text{for} \, t \in (t_0, T]. \label{eq:fast_cde_integral}
\end{align}
Here, $s$ denotes the integration variable. 
\revtext{Following the initialization strategy of Kidger et al.~\cite{kidger_neural_2020},} 
the initial states $\mathbf{d}(\tau_0)$ and $\mathbf{z}(t_0)$ are obtained from learned mappings: $\mathbf{d}(\tau_0) = \zeta_\phi(\mathbf{x}(\tau_0), \mathbf{u}(\tau_0), \tau_0)$ and $\mathbf{z}(t_0) = \xi_\theta(\mathbf{x}(t_0), \mathbf{u}(t_0), \mathbf{d}(t_0), t_0)$, where $\zeta_\phi: \mathbb{R}^{n+p+1} \rightarrow \mathbb{R}^{m}$ and $\xi_\theta: \mathbb{R}^{n+p+m+1} \rightarrow \mathbb{R}^{n}$.
By construction, $\tau_0 \ll t_0$, such that the slow CDE spans a longer integration horizon, while the fast CDE resolves shorter-term dynamics.
\revtext{The corresponding vector fields of the slow and fast CDEs, $g_\phi$ and $f_\theta$, are parametrized as Multi-Layer Perceptrons (MLPs).}
Both CDEs integrate over control paths $\mathbf{Y}(\tau)$ and $\hat{\mathbf{X}}(t)$. \revtext{Here, the degradation control path} $\mathbf{Y}(\tau)$ is generated via a path transformation \revtext{$h_{\phi}$, from the sensor measurements $\mathbf{X}(\tau)$ and exogenous inputs $\mathbf{U}(\tau)$, while the operational control path} $\hat{\mathbf{X}}(t)$ incorporates the interpolated slow state $\mathbf{d}(\tau(t))$ obtained from Eq.\eqref{eq:slow_cde_integral} \revtext{together with $\mathbf{X}(t)$ and $\mathbf{U}(t)$}. The detailed construction \revtext{of these control paths is provided} in Sec.~\ref{ssec:control_paths}.

Solving these CDEs involves two aspects of multiscale: multi-rate sampling of the data and multi-step numerical integration for each timescale. As shown in Fig.~\ref{fig:framework}, the slow CDE Eq.~\eqref{eq:slow_cde_integral} operates on coarsely sampled data to capture gradual degradation dynamics $g_\phi$, while the fast CDE Eq.~\eqref{eq:fast_cde_integral} uses finely sampled data to model rapid operational dynamics $f_\theta$.  
The construction of these data sequences will be detailed in Sec.~\ref{sec:method_training}.
This separation allows the use of multiscale numerical step sizes, larger for long horizon degradation accumulation and smaller for rapid operational fluctuations, which reduces stiffness and improves training efficiency. 

Finally, we introduce a readout mapping $\chi_{\theta}: \mathbb{R}^m \to \mathbb{R}^n$ that predicts the next system state $\hat{\mathbf{x}}(t+1)$ from the latent fast state $\mathbf{z}(t)$.

\subsection{Degradation Path Transformation and Construction}
\label{ssec:control_paths}
We first define the slow control path $\mathbf{Y}(\tau)$, which drives the degradation dynamics $g_\phi$ in Eq.~\eqref{eq:slow_cde_integral}. Direct interpolation of raw observations $(\mathbf{z}_i, \mathbf{u}_i, \tau_i)$, as commonly done in standard CDEs (Sec.~\ref {ssec:ncde}), is insufficient here since these fast states and inputs do not directly capture latent physical processes, such as stress, fatigue, or wear, that govern degradation rates. Moreover, the observed states evolve on a much faster timescale than degradation.

To address this, we introduce a \textbf{Degradation Path Transformation}. A learned mapping $h_\psi$ transforms observed states and \revtext{exogenous} inputs at each slow-time $\tau_i$ into degradation-relevant features:
\begin{equation}
    \mathbf{y}_i = h_\psi(\mathbf{x}_i, \mathbf{u}_i, \tau_i) \quad \text{for} \, \tau_i \in [\tau_0, T].
    \label{eq:path_transform_nn}
\end{equation}
These transformed features ${y}_i$ are then interpolated with a smooth interpolation scheme (e.g. natural cubic splines) over the sequence  $\{(\mathbf{y}_i, \tau_i) \mid \tau_i \in [\tau_0, T]\}$ to construct a continuous path $\mathbf{Y}(\tau)$. This differentiable trajectory serves as input to the fast CDE, enabling operation dynamics modeling that is both driven by \revtext{exogenous} inputs and degradation-aware. 

Next, we construct the fast control path $\hat{\mathbf{X}}(t)$, which drives the operational dynamics $f_\theta$ in Eq.~\eqref{eq:fast_cde_integral}. Unlike the slow control path $\mathbf{Y}(\tau)$, which requires a path transformation due to unobserved degradation drivers, the fast dynamics is captured directly from observable system states $\bm x(t)$, \revtext{exogenous} inputs $\mathbf{u}(t)$, and the interpolated degradation state $\bm d(\tau(t))$. Accordingly, $\hat{\mathbf{X}}(t)$ is obtained by interpolating $\{[\mathbf{x}_i, \mathbf{u}_i, \mathbf{d}(\tau(t_i)), t_i]\mid \tau_i \in [t_0, T] \}$. where $\mathbf{d}(\tau(t_i))$ aligns with the slow degradation trajectory at the fast-time index $t_i$. The resulting continuous path serves as the control path for the fast CDE, capturing fast system behavior conditioned on long-term degradation.

\subsection{Monotonicity Constraint for Degradation Dynamics}
\label{ssec:monotonicity}
Degradation in mechanical and structural systems is commonly modeled as an irreversible accumulation process. We therefore impose monotonicity on the inferred latent degradation state as an inductive bias~\cite{fink_physics_2026}, constraining the learned degradation dynamics to reflect this irreversibility and to suppress spurious variations induced by operational noise. 
This choice is also consistent with downstream RUL prediction, where monotonicity is widely regarded as essential for physical plausibility, identifiability, and stable long-term extrapolation~\cite{saxena_metrics_2008}.
To ensure our model respects this property and remains numerically stable during integration, the degradation rate $g_{\phi}d \bm Y$ in Eq.~\eqref{eq:fast_cde_integral} should ideally satisfy the following three conditions:
\begin{enumerate}
    \item \textbf{Non-negativity:} $\ge 0$, enforcing the monotonic growth of degradation.
    \item \textbf{Boundedness:} The rate must remain within a limited range to prevent excessively large growth during long horizon integration.
    \item \textbf{Zero response:} $g_\phi(\cdot)d\bm Y=0$, whenever $d\mathbf{Y}=0$ (no driving) or $g_\phi(\cdot)=0$ (no sensitivity), ensuring no change in degradation.
\end{enumerate}

\revtext{These requirements} can be achieved by applying an activation function to $g_{\phi}d \bm Y$. Standard activation functions fail to meet all these criteria simultaneously (e.g., ReLU/Softplus are unbounded; Sigmoid does not preserve zero response). \revtext{To address these limitations,} we propose a new activation function \revtext{that fulfills} these requirements:
\begin{equation}
    \sigma(x) = \text{sigmoid} (\gamma x) \cdot \tanh(x),
    \label{eq:pos_act}
\end{equation}
where $\text{sigmoid} (\cdot)$ is the sigmoid function, $\tanh(\cdot)$ is the hyperbolic tangent, and $\gamma \gg 1$ is a hyperparameter. \revtext{The intuition behind $\sigma(x)$ is to leverage the $\text{sigmoid}(\gamma x)$ term as a "soft step" gating mechanism that suppresses the negative quadrant of the $\tanh(x)$ term. By setting $\gamma \gg 0$, the gating term approaches zero rapidly for $x < 0$, ensuring the function is asymptotically non-negative. \revtext{For $x > 0$, the function behaves like $\text{tanh}(x)$, providing a strictly bounded derivative within $[0, 1)$.}
As a result, $\sigma(x)$ is strictly bounded and satisfies the zero-response condition} $\sigma(0) = 0$. 
While $\sigma(x)$ is not strictly non-negative, \revtext{due to a narrow region for $x\ll0$ where $\tanh(x)<0$ and $\text{sigmoid}(\gamma x)>0$, the magnitude of this effect is negligible for sufficiently large} $\gamma$. \revtext{For example, with $\gamma=10$, $\sigma(-1)\approx -3.4\times10^{-5}$. Increasing $\gamma$ further sharpens the gating effect and suppresses this residual negativity.}


\subsection{Model Training and Degradation Inference}
\label{sec:method_training}
The data are sampled at two rates. A \textit{long coarse sequence} $\{(\mathbf{x}_i, \mathbf{u}_i, t'_i)\}_{i=0}^N$, with a large sampling interval $\Delta t_s$ captures slow degradation dynamics over $[T_s, T]$, while a \textit{short dense sequence} $\{(\mathbf{x}_j, \mathbf{u}_j, t_j)\}_{j=0}^M$ with a small sampling interval $\Delta t_f$,  captures rapid operational dynamics over $[T_f, T]$. 
Both sequences end at the time $T$ ($t_M = t'_N = T$), but differ in resolution ($\Delta t_s \gg \Delta t_f$), sequence length ($N \gg M$), and starting times ($T_s \ll T_f$). 
Based on the observation up to time $T$, the model predicts the next fast state $\hat{\mathbf{x}}(t_{M+1})$ with $t_{M+1}=t_{M}+\Delta t_f$.

Model parameters are optimized by minimizing a multi-step rollout loss $\mathcal{L}$. 
Given history up to time $T=t_M$, the H-CDE model predicts $\hat{\mathbf{x}}(t_j)$ for $j=1, \dots, M+1$, \revtext{which is obtained by mapping the corresponding fast latent state $\mathbf{z}(t_j)$ through a readout layer parametrized by an MLP,} and $\mathcal{L}$ measures the discrepancy with ground truth observations $\mathbf{x}_j$:
\begin{equation}
    \mathcal{L}(\theta, \phi, \psi, \dots) = \frac{1}{M+1} \sum_{j=1}^{M+1} || \mathbf{x}_{j} - \hat{\mathbf{x}}(t_{j}) ||^2.
    \label{eq:training_loss_forecasting}
\end{equation}

By learning to forecast the observable fast states, the model implicitly captures the latent degradation dynamics $g_\phi$ and path transformation $h_\psi$, yielding the inferred degradation trajectory $\hat{\mathbf{d}}(\tau)$.

\section{Case Studies}
\label{sec:case-studies}
To demonstrate the effectiveness and adaptability of the proposed Hierarchical CDE (H-CDE) framework, we evaluate its performance on two distinct case studies: a simulated bridge undergoing stiffness loss due to dynamic traffic and thermal loading (Sec.~\ref{ssec:bridge-case}), and the benchmark N-CMAPSS turbofan engine dataset, where performance degrades in flow and efficiency (Sec.~\ref{ssec:cmapss-case}).. These systems were specifically chosen as they represent different underlying degradation processes and dynamical characteristics, with the bridge exhibiting dynamic response behavior and the engine operating under steady-state conditions, enabling a comprehensive evaluation of H-CDE across diverse domains.


\subsection{Degrading Bridge as a Damped Euler–Bernoulli Beam}
\label{ssec:bridge-case}
In the first case study, we model a simplified wooden bridge as a two-dimensional, simply supported Euler–Bernoulli beam with realistic material and geometric properties (detailed in \ref{app:bridge_params}). Degradation, modeled as a loss of material stiffness, is initiated when the beam’s vertical displacement exceeds a defined threshold. The bridge is subjected to a uniformly distributed vertical load representing railway traffic and a thermal load representing thermal expansion and bending due to ambient temperature variations.

\subsubsection{Dynamics Modeling} 
The bridge dynamics are governed by the following equation of motion 
\begin{align}
\bm M\,\ddot{\bm x}(t) + C\,\dot{\bm x}(t) + \bm K\,\bm x(t) = \bm f(t), \label{eq:bridge_system}
\end{align}
where $\bm x(t)$ represents the nodal degrees of freedom (axial displacement, vertical displacement, and rotation), and $\bm M, \bm C, \bm K$ are the global mass, damping, and stiffness matrices, respectively. We use the Newmark--beta method with sub-stepping to solve the structural dynamics. Damping follows the Rayleigh model, $\bm C = \alpha_{\text{damp}}\,\bm M + \beta_{\text{damp}}\,\bm K$, with coefficients set to achieve moderate damping ($\approx 4-5\%$ for the fundamental vibration mode). Specific parameter values are listed in \ref{app:bridge_params}.


\subsubsection{Load Modeling} 
The load vector $\bm f(t)$ includes the uniformly distributed vertical load $q(t)$ and the thermal loading $\bm{f}_T(t)$. The vertical load $\bm q(t)$ reflects realistic railway traffic load patterns based on statistical data, while the thermal load $\bm{f}_T(t)$ accounts both for thermal expansion from overall temperature changes and thermal bending due to temperature gradients across the beam depth, derived from real meteorological temperature measurements. Details on input load and temperature profile generation, as well as thermal load computation, are provided in \ref{app:bridge_params}.

\subsubsection{Degradation Modeling}
Structural degradation is captured by a scalar damage variable $D\in [0,1]$, where \(D = 0\) corresponds to an undamaged state and \(D = 1\) indicates complete stiffness loss. At each time step, we compute the absolute vertical displacement $v_{max}$ at mid-span $v_{\max} = \max_{\text{mid-span}} \bigl|v(t)\bigr|$. If $v_{max}$ exceeds a reference threshold $U_{\mathrm{ref}}$, the damage increment is given by $\Delta D = \beta_{\text{damage}} \, (1-D) \left( (v_{\max} - U_{\mathrm{ref}}) / U_{\mathrm{ref}} \right)^p$; otherwise, $\Delta D = 0$. The accumulated damage $D$ reduces the effective material stiffness through $E_{\mathrm{eff}} = E\,(1-D)$ and $I_{\mathrm{eff}} = I\,(1-D)$. The parameters governing damage evolution ($U_{\mathrm{ref}}, \beta_{damage}, p$) are provided in \ref{app:bridge_params}.

\subsubsection{Dataset Generation and Split}
\label{ssec:bridge_split}
We generated 12 bridge run-to-failure trajectories, each simulated until approximately 30\% stiffness reduction. Due to computational constraints, the material properties were chosen such that the simulated bridges would fail after about 2.5 months of operation, which also ensured larger environmental variations across different units.
The simulations cover two operational scenarios, both based on realistic railway traffic patterns and real ambient temperature data from different years, designed to evaluate generalization and out-of-distribution (OOD) performance.
Scenario A represents moderate daily load peaks with real meteorological data from one year, yielding five training trajectories and one in-distribution test trajectory. Scenario B introduces sharper daily load peaks with temperature data from a different year, yielding six OOD test trajectories. The increased load variations in Scenario B led to different dynamic responses, making it a challenging test case. 
Each trajectory simulates approximately two months of operational life. 
For downstream learning tasks, we extracted time series samples every 10 minutes, including displacements at quarter-, third-, and mid-span locations, the applied external load, and the ambient temperature. The complete data generation process is described in \ref{app:bridge_params}.

\subsection{Turbofan Engine Degradation (N-CMAPSS)}
\label{ssec:cmapss-case}
\subsubsection{Dataset Description and Split}
For the second case study, we utilize the NASA New Commercial Modular Aero-Propulsion System Simulation (N-CMAPSS) dataset \cite{chao_aircraft_2021}, which simulates turbofan engine degradation under realistic flight conditions. We focus on subset DS01, which involves a single component deterioration mode: High-Pressure Turbine (HPT) efficiency modifier. This enables us to treat the HPT efficiency modifier to serve as the ground-truth degradation state, in contrast to other subsets where multiple degradation mechanisms are present. 

To simplify evaluation, we consider only trajectory segments after the fault onset, corresponding to the exponential degradation phase, while excluding the initial linear progression and transition region. 
Additionally, since accelerated degradation is modeled at the flight level and held constant within each flight (i.e., independent of intra-flight operating conditions), the data represents degradation as discrete per-cycle updates rather than continuous degradation dynamics. Since the operating profile of a cycle differs across flight classes in both duration and load, these updates are not generalizable across classes.
To avoid this dataset-induced complication, we restrict our analysis to a single flight class.
Specifically, we focus on short-flight trajectories in DS01, providing a controlled setting for evaluating the model’s ability to disentangle slow degradation from fast operational variations.

We adopt a standard unit-based split; engine units 1, 4, and 7 are used for training, while Unit 9, which exhibits the fastest degradation among the four, is reserved for testing.

\subsubsection{Dynamics Modeling and Model Adaptation}
The N-CMAPSS model can be viewed as a discrete‐time dynamical system that reaches equilibrium at each operating point, producing steady-state engine outputs given the current inputs and health state~\cite{chao_aircraft_2021}:
\begin{align}
[\mathbf{x}_s(t), \mathbf{x}_v(t)] = f\big(\mathbf{u}(t), \theta(t)\big), \label{eq:ncmaps_system}
\end{align}
where \(\mathbf{u}(t)\) denotes \revtext{exogenous} inputs (e.g., altitude, Mach number, throttle angle, and inlet temperature), and \(\theta(t)\) contains health-related parameters (e.g., the HPT-efficiency modifier). 
The model outputs include observable physical properties \(\mathbf{x}_s(t)\) and latent internal states \(\mathbf{x}_v(t)\), the latter serving as virtual sensors not accessible through direct measurements.
Following fault onset, accelerated degradation is modeled by updating \(\theta(t)\) between flight cycles via a stochastic exponential decay applied to component flow capacities and efficiencies.

To align with this system dynamics, we adapt the fast dynamics module of the H-CDE (Eq. \ref{eq:fast_cde_integral}) by removing the dependence of $d\mathbf{x}/dt$ on $\mathbf{x}(t)$, since the N-CMAPSS outputs are generated directly from inputs and the degradation state. We modify the inputs to $f_\theta$ in Eq. \ref{eq:fast_cde_integral}, making $\mathbf{u}(t)$ and $\mathbf{d}(t)$ the primary drivers for $d\mathbf{x}(t)/dt$ via the \revtext{degradation} control path \revtext{$\mathbf{Y}(t)$}.

\section{Experimental Setup}
\label{sec:exp_setup}
\revtext{This work primarily addresses scenarios where system degradation is inherently unobservable or prohibitively expensive to measure directly. 
Thus, we design the degradation as an unsupervised task where the goal is to infer the degradation progression from the sensor data and only use the true degradation label for evaluation.}
This section outlines the experimental design used to assess the quality of inferred latent degradation states by the proposed method. We describe the setup and configuration of the residual-based baseline in Sec.~\ref{ssec:exp_baseline}, define the evaluation metrics for latent degradation quality in Sec.~\ref{ssec:exp_baseline}, and provide details on training setups, data preprocessing configurations, hyperparameters, as well as hardware specifications in Sec.~\ref{ssec:implementation_details}.
\revtext{Furthermore, to evaluate the structural effectiveness of the hierarchical architecture independently of the unsupervised constraint, we provide a comparison with state-of-the-art supervised models in \ref{sec:eval_supervised_task}.}

\subsection{Baseline: Residual-based Model}
\label{ssec:exp_baseline}
We compare H-CDE against a residual-based baseline that infers degradation indirectly by measuring deviations from a model of healthy system behavior. Specifically, we adopt the operating-condition-based residual model from~\cite{hsu_comparison_2023}, which is widely used and has demonstrated strong empirical performance.
This involves training a Feedforward Neural Network (FNN) exclusively on data from the healthy operational phase to learn the mapping from \revtext{exogenous} inputs $\mathbf{u}$ to operating states $\mathbf{x}$. Depending on the nature of system dynamics (i.e, steady-state vs dynamic response systems), the model may also incorporate a history of past states $\mathbf{x}$ as additional input. The prediction error (residual) on new data then serves as a degradation indicator. 

For the bridge case study, where the system dynamics explicitly depend on the current state and no purely healthy samples are available, we approximate healthy behavior using the first 1000 samples (approximately 10\% of each trajectory). We train an FNN to predict the current state $\hat{\mathbf{x}}_k$ from a flattened sequence of the past $w-1$ states $\{\mathbf{x}_j\}_{j=k-w+1}^{k-1}$ and the sequence of $w$ \revtext{exogenous} inputs $\{\mathbf{u}_j\}_{j=k-w+1}^{k}$. 
For the N-CMAPSS case study, where system dynamics are primarily driven by \revtext{exogenous} inputs and healthy cycles are labeled, we train an FNN on healthy data to learn a direct mapping from current control input $\mathbf{u}_k$ to the current state $\hat{\mathbf{x}}_k$. 


\subsection{Evaluation Metric} 
\label{ssec:metric}
To assess the quality of the learned latent degradation state, we introduce an \textbf{Alignment Score} that measures its linear correlation with the true degradation. Prior works typically focus on health index construction, where degradation is represented as a scalar value and evaluated using dedicated metrics. These metrics, however, are not directly applicable in our setting, since the extracted degradation states can be high-dimensional. The core idea of the proposed metric is that a well-disentangled representation should allow the true degradation to be recovered through a simple linear transformation. 
Concretely, we train a linear regression model on the latent representations from the training set to predict the true degradation, and then apply the learned regressor to the test set. The \textbf{Alignment Score} is defined as the coefficient of determination ($R^2$) between the predicted and true degradation values on the test set. \revtext{In addition, to analyze the computational efficiency of different H-CDE architectural variants in the ablation studies, we report the \textbf{Number of Function Evaluations (NFE)}, which serves as a hardware-agnostic proxy for computational complexity.}

\subsection{Ablation Studies}
\label{ssec:exp_setup_abl}
To assess the contribution of individual components in our proposed model, we evaluate the following variants in both case studies:
\begin{itemize}[leftmargin=1.5em, itemsep=4pt, topsep=4pt, partopsep=0pt, parsep=0pt]
    \item \textbf{w/o MC}: Removes the monotonic constraint on the learned degradation states, allowing non-monotonic degradation trajectories.
    \item \textbf{w/o PT}: Removes the path transformation applied to the input data prior to the CDE integration.
\end{itemize}


\subsection{Implementation Details}
\label{ssec:implementation_details}

\subsubsection{Data Processing Configuration}
The bridge dataset is sampled at 1~Hz, while the N-CMAPSS dataset was downsampled to 0.1~Hz following the recommendation of Arias Chao et al.~\cite{chao_aircraft_2021}.
For the residual model, we performed a sweep over input window sizes \( w \in \{3, 5, 10, 20\} \). The best performance on the bridge dataset was achieved with \( w = 5 \). For N-CMAPSS, we set \( w = 5 \), consistent with its steady-state system dynamics and prior work~\cite{hsu_comparison_2023}. Additionally, we tuned the healthy data hold-out length, exploring values of 500, 1000, and 2000, and selected 1000 as optimal.

For H-CDE, inputs are constructed as multiscale sequences. 
In the bridge case study, we use a slow sequence of length $w_s=100$ with time step $\Delta t_s = 12$ and a fast sequence of length $w_f=11$ with $\Delta t_f = 1$, both generated with a sliding window stride of two. The slow sequence length $w_s=100$ was fixed, and $\Delta t_s = 12$ was chosen to span approximately 10\% of the full run-to-failure trajectory. The fast sequence length $w_f=11$ was selected to align with the final slow step time interval. The stride was set to two to reduce the volume of training data. 
In the N-CMAPSS case study, the slow sequence also uses $w_s=100$, but with $\Delta t_s = 20$, and the fast sequence uses $w_f=5$ with $\Delta t_f = 2$, generated with a stride of five. The slow sequence configuration was chosen to roughly span three short flight cycles. For fast sequence length $w_f$, we compared lengths of five and ten, and selected \( w_f = 5 \) based on better empirical performance.

\subsubsection{\revtext{Architectural Specifications and Hyperparameters}}
\textbf{Residual} (Bridge: 4,963 params; N-CMAPSS: 4,184 params).
We employ a Feedforward Neural Network (FNN) with 4 hidden layers, using SiLU activation, batch normalization, and a dropout rate of 0.2. The hidden layer dimensions were selected from the same search space  \{[20,20,20,10], [100,100,50,20], [50,50,20,10]\}, with [50, 50, 20, 10] yielding the best performance across both datasets.

\textbf{H-CDE} (Bridge: 23,268 params; N-CMAPSS: 16,636 params). 
To reduce the hyperparameter search space, we applied the same configuration to both slow and fast CDE layers. Vector fields \revtext{$g_\phi(\cdot)$ and $f_\phi(\cdot)$ in both CDE layers}, as well as \revtext{the path transformation $h_\psi$ and} the readout head\revtext{, are implemented as} 2-layer MLPs. We explored \revtext{MLP} hidden dimensions of 64 and 128 and found 64 to yield better performance for both case studies. Similarly, we tuned the latent dimensions \revtext{of both CDE layers} between 10 and 20, with 10 yielding the best results in both cases.
For the N-CMAPSS dataset, the latent dimension of slow CDE was further tuned between five and ten, with five selected as optimal.
SiLU activation was used, without batch normalization or dropout.
Following common practice for NODE-based models, no explicit initialization was applied due to sensitivity to initial conditions. Both case studies used an explicit adaptive solver dopri5 with relative tolerance $1\text{e}^{-3}$ and absolute tolerance $1\text{e}^{-5}$.

\subsubsection{Training Setups}
\label{sec:common_training_params} 
All models were trained using the AdamW optimizer~\cite{loshchilov_decoupled_2019} with an initial learning rate of $5 \times 10^{-3}$ and a batch size of 256. The models minimize the L2 (Mean Squared Error) loss. From the generated training samples, 20\% are randomly held out as a validation set. This validation set is used for early stopping decisions and for learning rate scheduling via ReduceLROnPlateau, which reduces the learning rate by a factor of 0.95 if the validation loss does not improve for five consecutive epochs.
For the bridge dataset, training proceeds for a maximum of 30 epochs with early stopping and a minimum of five epochs. 
For the N-CMAPSS dataset, training runs for a maximum of 60 epochs and a minimum of 20 epochs.

\subsubsection{Hardware and Software}
\label{sec:hardware_software} 
All methods were implemented using PyTorch 1.12.1~\cite{paszke_pytorch_2019} with CUDA 12.0. Computations for both case studies were executed on a GPU cluster equipped with NVIDIA A100 80GB GPUs. We employed neptune.ai for experiment tracking and management.

\section{Results and Discussions}
\label{sec:results}
This section presents our experimental results on self-supervised degradation inference. We evaluate the quality of the learned latent representation on two case studies: the \textbf{Bridge} dataset, which represents a system with dynamic response (Sec.~\ref{ssec:result_bridge}), and the \textbf{N-CMAPSS} dataset, which represents a steady-state system (Sec.~\ref{ssec:result_cmapss}). Relevant ablation studies analyzing the contribution of individual model components are discussed within each subsection.


\subsection{Bridge Case Study}
\label{ssec:result_bridge}
\begin{figure}[tbh]
  \centering
    \includegraphics[width=1\linewidth]{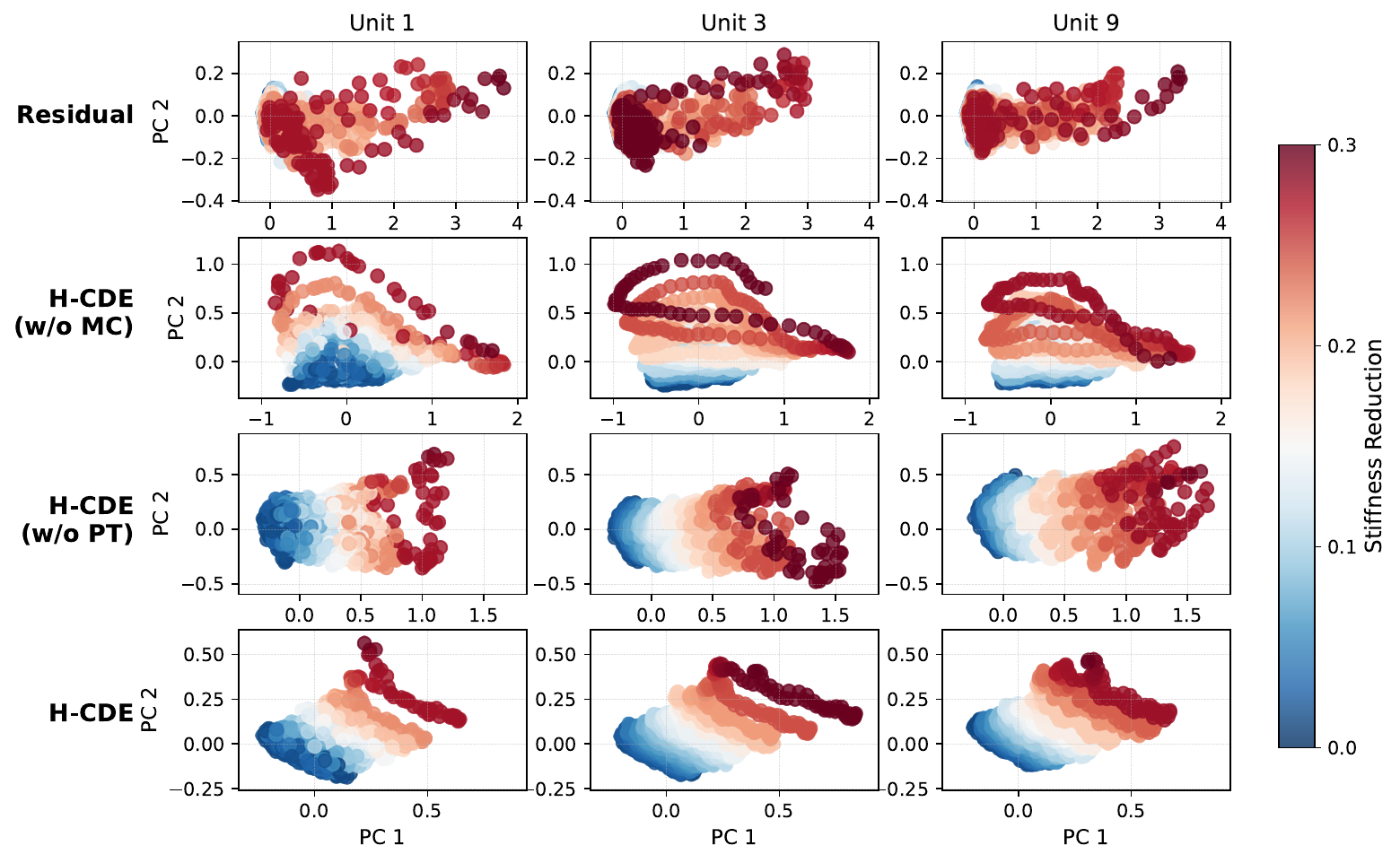}
 \caption{First two principal components (PCs) of embedding space for selected units in the bridge dataset (Training Unit 1, ID Test Unit 3, OOD Test Unit 9). Colors indicate true degradation (stiffness reduction).} 
  \label{fig:bridge_latent}
\end{figure}

We first examine the embedding space of the residuals from the residual method and the latent degradation states from H-CDE and its variants on the bridge dataset, as shown in Fig.~\ref{fig:bridge_latent}. 
For the residual method, all residuals are reduced to two dimensions using PCA for visualization purposes, and the latent states from H-CDE and its variants are projected in the same way. The plots show the first two principal components (PCs), for representative training (unit 1), in-distribution (ID) test (unit 3), and out-of-distribution (OOD) test units (unit 9). 
The first two units share the same load distribution, whereas the last differs.
The specific train–test split was described in Sec.~\ref{ssec:bridge_split}.
To quantify the quality of these embedding spaces, we report the alignment score as defined in Sec.~\ref{ssec:metric} in Tab.~\ref{tab:alignment_scores}. 
A detailed comparison of the extracted latent degradation representations is presented in Sec.~\ref{ssec:res_bridge_embed}, followed by the ablation study results in Sec.~\ref{ssec:res_bridge_abl}, which analyze the impact of the monotonicity constraint, path transformation, and slow time step size on both alignment score and the computational cost.

\begin{table}[tbh]
\centering
\footnotesize
\caption{Alignment Score ($R^2$) on in-/out-of-distribution runs.}
\label{tab:alignment_scores}
\begin{tabular}{lcc}
\toprule
Method & In-Distribution & Out-of-Distribution \\
\midrule
Residual        & 0.324 ± 0.054 & 0.282 ± 0.036 \\
H-CDE (w/o MC)  & 0.969 ± 0.010 & 0.941 ± 0.033 \\
H-CDE (w/o PT)  & 0.954 ± 0.010 & 0.820 ± 0.045 \\
H-CDE           & 0.967 ± 0.007 & 0.941 ± 0.018 \\
\bottomrule
\end{tabular}
\end{table}

\subsubsection{Comparison of Latent Degradation Representation}
\label{ssec:res_bridge_embed}
\textbf{Limitations of the residual-based method}.
As shown in Fig.~\ref{fig:bridge_latent}, the embedding space derived from the residuals provides limited information about the true degradation.
This is consistent with the low alignment score reported in Tab.~\ref{tab:alignment_scores}.
Although the FNN with past state inputs can model healthy system dynamics reasonably well, the residuals do not directly represent the degradation state. 
In the literature, residuals, or their distributions, are often used as a proxy for degradation level~\cite{sarwar_probabilistic_2024}. However, this approach does not account for the dynamic responses, and the residuals therefore fail to capture the nonlinear effects of degradation on sensor measurements, leaving these effects entangled with the load and temperature history.

\textbf{H-CDE: Meaningful degradation embedding and generalization}. 
Compared with the residual baseline, the H-CDE embeddings in Fig.~\ref{fig:bridge_latent} form a rectangular manifold on which the degradation trend is clearly visible, indicating successful disentanglement and physically meaningful latent states.  The enlarged spacing toward the end of the trajectory suggests an accelerating degradation rate as the system approaches the end of operational life, consistent with the expected accelerated stiffness loss in the underlying physical model.
The embeddings are consistent across training, in-distribution test, and OOD test units, which aligns with the results in Tab.~\ref{tab:alignment_scores}, where H-CDE achieves the highest alignment score for both cases. This consistency demonstrates strong generalization to varying and unseen operating and environmental conditions.
A notable observation is that, although PC1 explains most of the variance, the degradation trajectory is inclined in the (PC1, PC2) plane.
We hypothesize that this tilt arises from modeling the underlying second-order damped dynamics (Eq.\eqref{eq:bridge_system}) with a first-order CDE for the fast dynamics (Eq.\eqref{eq:fast_cde_integral}). Unmodeled second-order dynamics may introduce additional first-order operational variations into the latent states, leading to this oblique manifold.
While accurately modeling the fast system dynamics is beyond the scope of this work, which focuses on disentanglement, future research could explore higher-order NODEs~\cite{massaroli_dissecting_2020} into the fast CDE to better capture these dynamics and potentially improve performance.

\subsubsection{Ablation Studies and Sensitivity Analysis}
\label{ssec:res_bridge_abl}
\begin{table}[tbh]
\caption{Ablation study of H-CDE: slow CDE number of function evaluations (NFE).}
\label{tab:nfe}
\centering
\footnotesize
\begin{tabular}{lcc}
\toprule
Method & In-Distribution & Out-of-Distribution \\
\midrule
H-CDE           & 689 ± 5 & 631 ± 93 \\
H-CDE (w/o MC)  & 712 ± 11 & 996 ± 109 \\
H-CDE (w/o PT)  & 686 ± 1 & 748 ± 44 \\
\bottomrule
\end{tabular}
\end{table}
\textbf{Impact of monotonicity constraint}.
Although the alignment score of the w/o MC variant is close to that of H-CDE (Tab.~\ref{tab:alignment_scores}), its embedding space in Fig.~\ref{fig:bridge_latent} differs substantially.
Without the monotonicity constraint, the magnitude of the latent states no longer predominantly increases with the degradation state.
Furthermore, PC1 no longer aligns with the primary degradation trend, which instead appears along PC2, indicating that the embedding space is still dominated by operational variations.
Moreover, the embeddings for training and test units differ more noticeably, indicating reduced generalization.
This is also reflected in Tab.~\ref{tab:nfe}, where w/o MC requires significantly more function evaluations (NFE), particularly for OOD data (about 58\% higher), thereby increasing computational cost.
Overall, the monotonicity constraint yields more interpretable degradation representations and acts as a valuable regularizer, guiding optimization toward stable solutions with lower NFE and physically consistent degradation dynamics.


\begin{figure}[tbh]
  \centering
    \includegraphics[width=.9\linewidth]{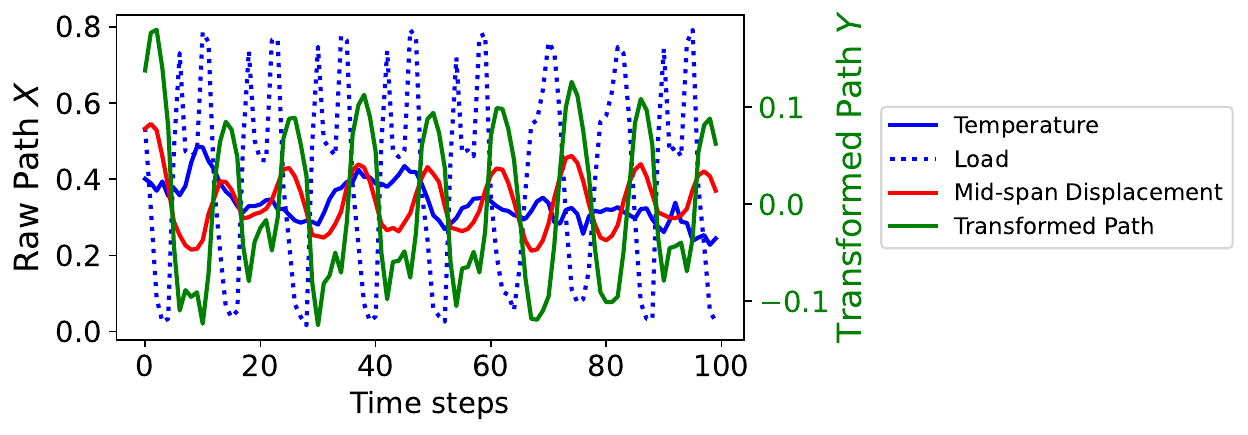}
\caption{Path transformation for a sample from the bridge dataset.}
  \label{fig:bridge_trans}
\end{figure}

\textbf{Impact of path transformation}.
Removing the path transformation (w/o PT) has a greater impact on the alignment score than removing the monotonicity constraint, particularly for the OOD subset (Tab.~\ref{tab:alignment_scores}).
In Fig.~\ref{fig:bridge_latent}, the overall embedding structure of w/o PT is similar to that of H-CDE, with PC1 capturing most of the degradation variance.
However, the shape suggests some unexplained variance from a third dimension, and the OOD embeddings differ more noticeably, which is consistent with the lower score in Tab.~\ref{tab:alignment_scores}.
\revtext{The path transformation is a learnable temporal mapping that converts raw exogenous inputs and sensor measurements into a multivariate latent path, designed to extract slow, degradation-relevant temporal structure. It is learned end-to-end from data and optimized for downstream degradation inference without requiring prior knowledge of or imposing an explicit physics-based degradation model.}
To illustrate \revtext{its} effect, Fig.~\ref{fig:bridge_trans} shows the raw input path for a sample (temperature, load, and displacement) alongside the learned transformed path.
The transformation originally produces a 10-dimension\revtext{al latent degradation representation}. Here, only its first principal component after performing PCA is \revtext{plotted for better visualization}.
Notably, this primary transformed path does not track variations in \revtext{exogenous inputs such as} load or temperature but more closely resembles the displacement signal, with narrower peak regions.
\revtext{While the transformed path should not be interpreted as an explicit physical degradation drivers such as stress or cyclic loading, its temporal structure reflects degradation-relevant behavior present in the inputs.
In this example, the emphasis on displacement-related temporal regions is qualitatively consistent with the simulated bridge degradation process, in which damage accumulates primarily when displacement exceeds a certain range. Accordingly, the transformed path exhibits low values during periods of small displacement, capturing when degradation progression is inactive rather than explicitly modeling physical quantities such as stress.}
\revtext{By extracting degradation-relevant features evolving on a slower time scale, rather than operating directly on raw inputs dominated by fast operating dynamics, the path transformation also improves numerical efficiency. As shown in Tab.~\ref{tab:nfe}, H-CDE with path transformation requires fewer} NFEs than the w/o PT variant in the OOD \revtext{setting, indicating improved numerical conditioning and generalization. For the ID setting, the difference in NFEs is small as the raw displacement signal already aligns well with the dominant degradation driver.}

\begin{figure}[tbh]
  \centering
    \includegraphics[width=.9\linewidth]{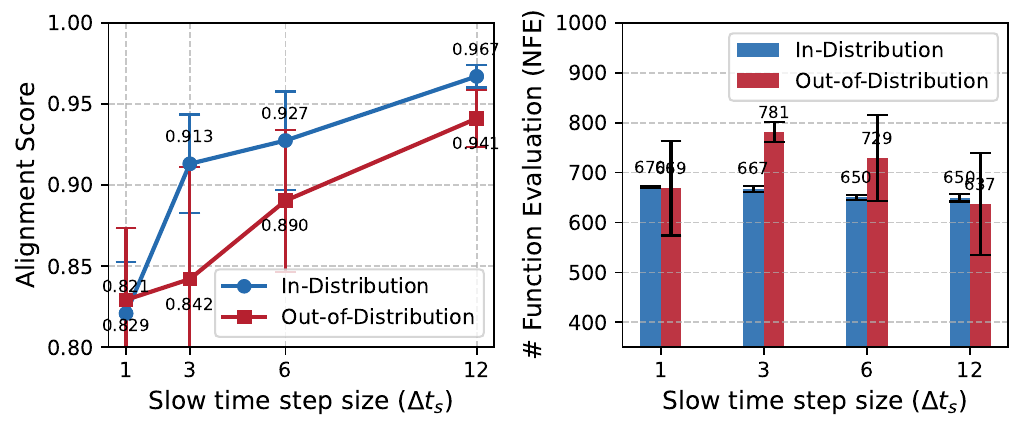}
     \caption{Sensitivity of H-CDE performance to the slow time step size ($\Delta t_s$) in the bridge case study.}
\label{fig:bridge_slowtimesize_ablation}
\end{figure}

\textbf{Impact of slow time step size}.
In this sensitivity analysis, we vary the slow time step size $\Delta t_s$ while keeping the fast time step size ($\Delta t_f = 1$) and the fast window size ($w_f = 100$) fixed, thereby changing the total time interval covered without altering the number of input samples.
We evaluate $\Delta t_s \in \{1, 3, 6, 12\}$, stopping at 12 as it corresponds to roughly 10\% of the service life; larger values would require more data and reduce usability.
The resulting alignment score and NFE are shown in Fig.~\ref{fig:bridge_slowtimesize_ablation}.
The alignment score generally increases with $\Delta t_s$ for both ID and OOD subsets, indicating that longer temporal contexts help capture degradation progression.
For ID subset, high alignment scores are already achieved at $\Delta t_s = 3$, while the OOD subset requires larger steps ($\Delta t_s \approx 6$) to reach comparable scores, suggesting that longer-range dependencies are more critical for generalization.
In terms of computational cost (NFE), ID subset remains stable across $\Delta t_s$, whereas OOD cost decreases when $\Delta t_s \ge 3$.
One possible explanation is that the coarser, ``rougher" path representation from larger steps is sufficient to capture the degradation progression while simplifying the dynamics for the adaptive ODE solver, reducing noise and complexity~\cite{morrill_neural_2021}, thereby improving generalization.
Overall, $\Delta t_s$ controls a trade-off between data requirements and the ability to better extract latent degradation trends.

\subsection{N-CMAPSS Case Study}
\label{ssec:result_cmapss}
\begin{figure}[tbh]
  \centering
    \includegraphics[width=.75\linewidth]{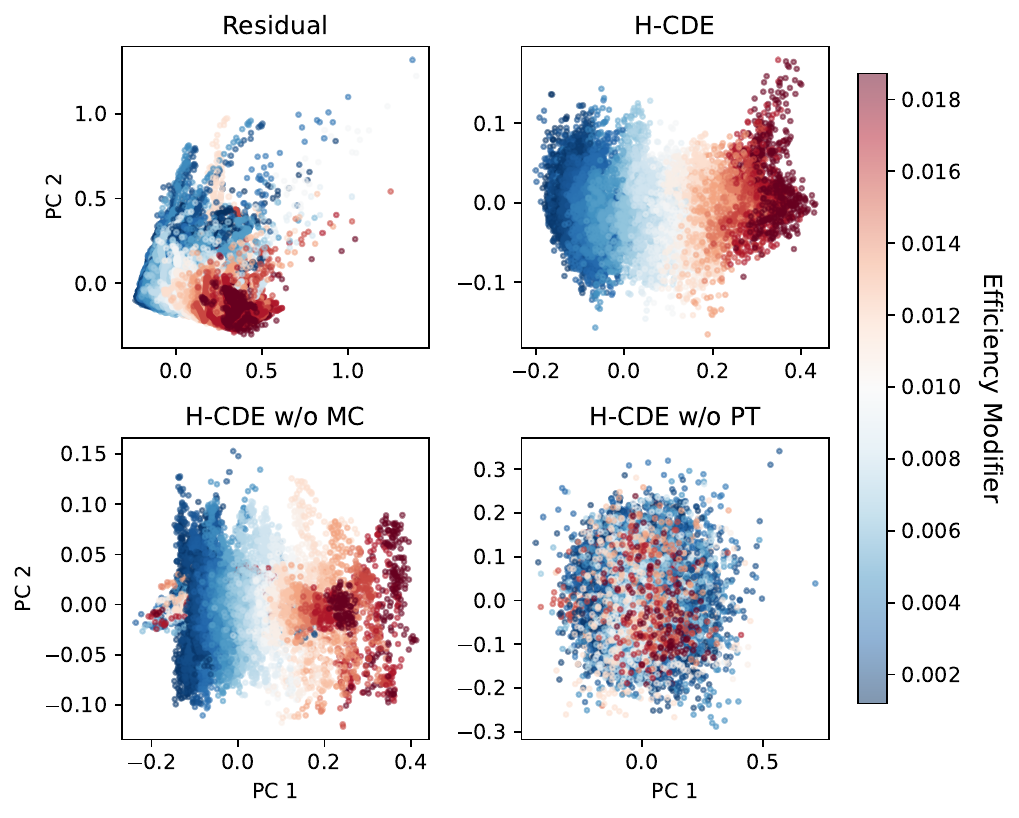}
    \caption{Latent degradation states of the test unit 9 from the N-CMAPSS dataset, projected onto PC1/PC2, shown across five random seeds.}
\label{fig:cmapss_embedding}
\end{figure}

\begin{figure}[tbh]
  \centering
    \includegraphics[width=.8\linewidth]{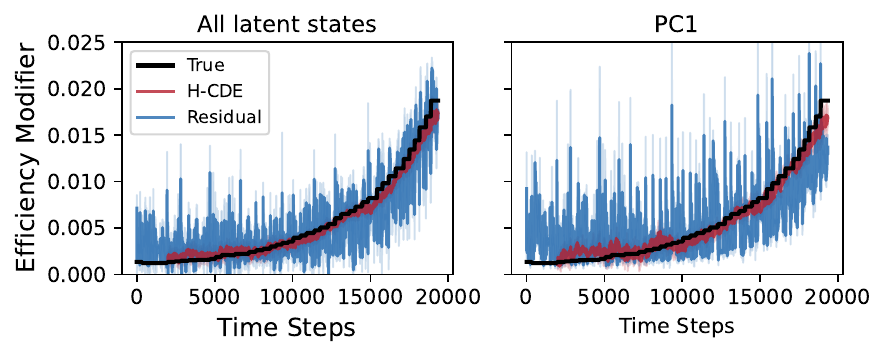}
  \caption{Predicted degradation (efficiency modifier) for N-CMAPSS using linear regression on H-CDE latent states versus baseline residuals. Left: predictions from all latent dimensions. Right: predictions using only the first principal component (PC1)}
  \label{fig:cmapss_eff}
\end{figure}

For the N-CMAPSS case study, we qualitatively and quantitatively analyze the embedding space of both residual-based methods and H-CDE, along with its variance, following the same procedure as in the bridge case (see Fig.~\ref{fig:cmapss_embedding}, Tab.~\ref{tab:cmapss_alignment}).
To assess the practical utility of the degradation embeddings, we evaluate the alignment score of the first principal component (PC1), from which a potential health index can be derived when full run-to-failure trajectories are available. Predicted degradation state (efficiency modifier) derived from either all latent dimensions or PC1 alone is shown in Fig.~\ref{fig:cmapss_eff}, with corresponding alignment scores reported in Tab.~\ref{tab:cmapss_alignment}. These results are discussed in Sec.~\ref{ssec:result_embed_cmapss}, with additional ablation studies in Sec.~\ref{ssec:res_cmapss_abl}.

\subsubsection{Comparison of Latent Degradation Representation}
\label{ssec:result_embed_cmapss}

\textbf{Residual baseline: entanglement with operating conditions}.
The embedding extracted by the residual method in Fig.~\ref{fig:cmapss_embedding} better reveals the degradation trend than in the bridge case, with PC1 roughly reflecting the degradation level in Fig.~\ref{fig:cmapss_eff}. 
However, it remains noisy, with fluctuations appearing to reflect operational cycles. When using only PC1, these fluctuations become more pronounced, as also seen in the lower alignment score in Tab.~\ref{tab:cmapss_alignment}. This indicates that the residual-based method cannot effectively disentangle degradation from operational variations.
This limitation arises from its assumption that sensor measurements under degradation can be expressed as a linear combination of healthy behavior and degradation effects, i.e., $\mathbf{r}_k = \mathbf{x}_k - \hat{\mathbf{x}}_k \approx g(\mathbf{u}_k, \mathbf{d}_k) - f(\mathbf{u}_k)$, 
which implies $\bm{x}_k = f(\mathbf{u}_k) + \mathbf{d}_k$.
In reality, degradation and \revtext{exogenous} inputs interact nonlinearly in the true system dynamics $g(\mathbf{u}_k, \mathbf{d}_k)$~\cite{si_nonlinear_2022}. These unmodeled nonlinear dependencies remain entangled with operating history, leading to noisy estimates.
Compared to the bridge case, the residual method performs better on N-CMAPSS since it is a steady-state system where $\mathbf{x}(t)$ depends directly on $\mathbf{u}(t)$ and $\mathbf{d}(t)$, without derivative terms. This reduces entanglement and yields clearer embeddings.

\textbf{H-CDE: accurate degradation representation.}
In comparison, the H-CDE embedding is much more structured, with PC1 capturing most of the degradation variance in Fig.~\ref{fig:cmapss_embedding}. 
This is also evident in Fig.~\ref{fig:cmapss_eff}, where the predicted degradation closely matches the ground truth, even when using only PC1. Unlike the residual baseline, the predictions are stable within operational cycles. This is consistent with the quantitative results in Tab.~\ref{tab:cmapss_alignment}, where H-CDE achieves about 20\% higher alignment score than the baseline when using all latent states/residuals, and an even larger improvement of around 64\% when using only PC1. These high alignment scores validate that the learned degradation representation generalizes well to the new test unit with different operating cycles. Finally, the comparable alignment scores from all states and from PC1 show that a simple dimensionality reduction method such as PCA can yield a scalar degradation representation, from which a health index can be derived when run-to-failure trajectories are available, highlighting the practical utility of H-CDE.

\subsubsection{Ablation Studies}
\label{ssec:res_cmapss_abl}
\begin{figure}[tbh]
  \centering
    \includegraphics[width=1\linewidth]{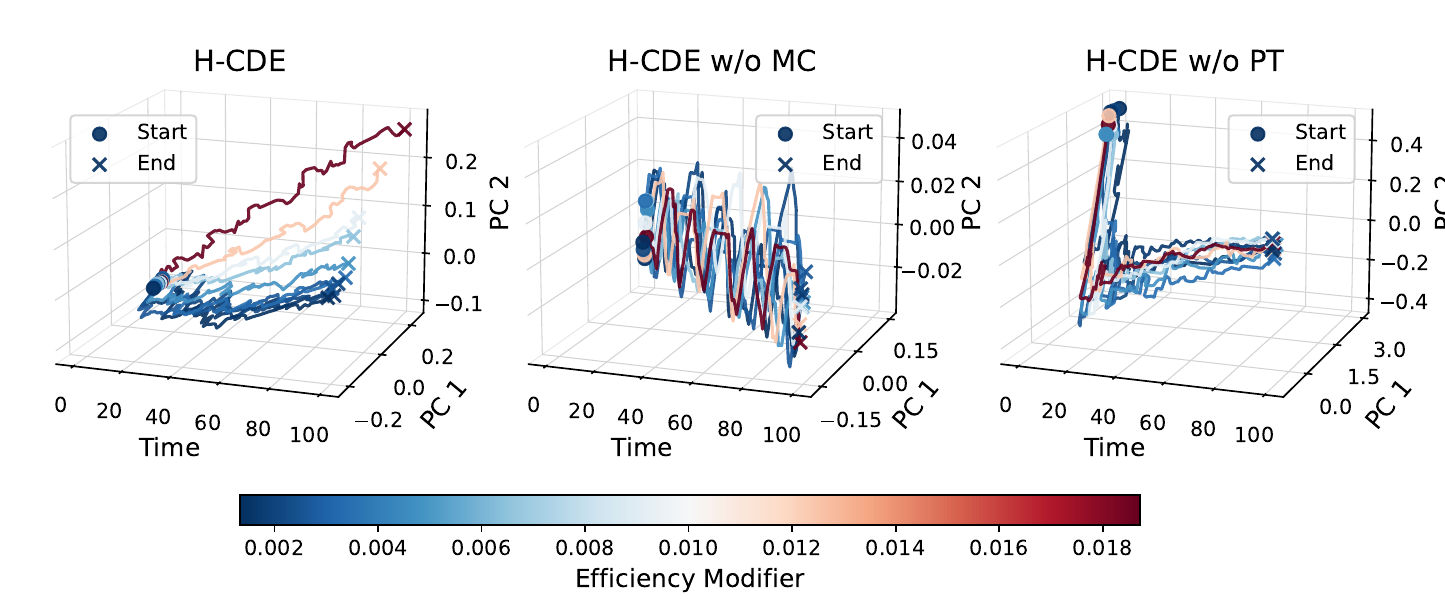}
    \caption{Degradation trajectories $\bm d(\tau)$ in the N-CMAPSS ablation study. Each trajectory represents one sample, colored by its final degradation level (efficiency modifier).}
\label{fig:cmapss_abl_trajs}
\end{figure}

We now examine the ablation studies on the N-CMAPSS dataset. Compared to the bridge case, the impact of path transformation and monotonicity constraint is much larger. To better understand this behavior, we additionally visualize the learned degradation trajectories $\bm d(\tau)$ in Fig.~\ref{fig:cmapss_abl_trajs}. Each trajectory corresponds to one data sample, colored by its final degradation level, with a sequence length matching the slow-sequence window size. The endpoint of each trajectory in Fig.~\ref{fig:cmapss_abl_trajs} corresponds to a point in the embedding space shown in Fig.~\ref{fig:cmapss_embedding}. The following summarizes the main insights from these ablation results.

\textbf{Degradation trajectories of H-CDE}.
For H-CDE, the trajectories exhibit clear separation according to the degradation level and progress monotonically along the primary degradation axis (PC1) over time. The rate of this progression varies based on the trajectory's degradation level: trajectories with higher degradation progress more rapidly, diverging from healthier ones and converging to distinct, separable endpoints in Fig.~\ref{fig:cmapss_abl_trajs}. Interestingly, the learned trajectories resemble exponential curves, consistent with the underlying physical degradation process.

\textbf{Impact of path transformation.}
Path transformation is found to be critical in this case study. The w/o PT variant achieves near-zero alignment scores in Tab.~\ref{tab:cmapss_alignment} and yields an uninterpretable embedding space in Fig.~\ref{fig:cmapss_embedding}. The latent trajectories in Fig.~\ref{fig:cmapss_abl_trajs} explain this deficiency: along the primary degradation axis (PC1), they first briefly decrease before increasing, breaking monotonic progression.
This occurs despite the MC constraint, due to the small negative range permitted by the activation function (see Eq.\ref{eq:pos_act}). As a result, trajectory endpoints collapse into a single region, indicating that the model fails to capture physically meaningful degradation paths and to separate degradation levels.
These findings highlight path transformation as a key mechanism for isolating degradation drivers from dominant operational dynamics and enabling the model to learn the degradation process. Its impact is notably stronger in N-CMAPSS than in the bridge case. This increased sensitivity can be attributed to two factors: (i) in the bridge case study, the main degradation-related state (displacement) is directly measured, reducing the need for path transformation; and (ii) N-CMAPSS has a higher-dimensional sensor space with hidden degradation drivers, making path transformation essential for isolating key degradation drivers.

\begin{table}[tbh]
\centering
\footnotesize
\caption{Alignment Score for Different Ablation Settings}
\label{tab:cmapss_alignment}
\begin{tabular}{lcc}
\toprule
       Model & All latent states & PC1 \\
\midrule
    Residual &          0.802 ± 0.024 &  0.566 ± 0.072 \\
CDE (w/o MC) &          0.820 ± 0.219 &  0.691 ± 0.374 \\
CDE (w/o PT) &          0.003 ± 0.064 & -0.026 ± 0.057 \\
         CDE &          0.964 ± 0.010 &  0.928 ± 0.028 \\
\bottomrule
\end{tabular}
\end{table}

\textbf{Impact of the monotonicity constraint.}
Removing the monotonicity constraint has a smaller impact than removing path transformation, but it still noticeably degrades the performance compared to the full H-CDE model. The effect is most evident when using only PC1, where alignment scores are lower and exhibit higher variance as reported in Tab.~\ref{tab:cmapss_alignment}. This instability reflects inconsistent separability in the learned latent space across runs, as shown in Fig.~\ref{fig:cmapss_embedding}.
This behavior can be explained by inspecting the latent trajectories in Fig.~\ref{fig:cmapss_abl_trajs}: in failed runs, the w/o MC variant exhibits cyclic patterns that resemble operational cycles, indicating a stronger influence from operational variations than degradation progression. Without the monotonicity constraint, the model struggles to enforce a clear degradation trend, leading trajectories to collapse toward a common region and reducing state separability.
This challenge is amplified in the N-CMAPSS case study because degradation remains constant within each operational cycle (a simulation limitation) and only changes between cycles. The strong cycle-to-cycle operational variation makes it difficult for the model to infer the slow underlying progression without explicit guidance. The monotonicity constraint therefore plays a crucial role in stabilizing training, improving embedding quality, and ensuring a physically consistent degradation trajectory.

\section{Conclusion \revtext{and Future Works}}
\label{sec:conclusion}
This paper proposes the Hierarchical Controlled Differential Equation (H-CDE) framework, a novel approach for disentangling degradation from operational dynamics in sensor data. H-CDE models degradation and operating states as coupled slow-fast dynamics, parameterized by separate Controlled Differential Equations (CDEs). Its key contributions include an efficient multi-scale integration scheme to mitigate numerical stiffness, a learnable path transformation to derive latent degradation drivers as the control path in the slow degradation CDE, and a monotonicity constraint in the degradation progression, implemented through a custom activation function that regularizes disentanglement.
We investigated two representative systems: bridges, which exhibit dynamic responses, and aero-engines, which operate in steady states, where most existing research has focused. Experiments demonstrated H-CDE’s ability to capture long-range dependencies and multi-scale dynamics, consistently outperforming residual-based baselines on both system types. Qualitative analysis of the learned degradation embeddings and quantitative alignment with true degradation states confirmed its effectiveness, while ablation studies highlighted the importance of the path transformation and monotonicity constraint. We further highlighted the limitations of widely used residual-based methods, which model \revtext{sensor measurements as} as a superposition of \revtext{healthy system behavior} and a \revtext{an additive} degradation\revtext{-induced deviation}. This assumption leads to noisy inference in steady-state systems and fails entirely in dynamic-response systems, where residuals are heavily entangled with operational states.

Overall, this work highlights the need to account for the specific nature of system dynamics in degradation inference. By jointly modeling operational and degradation dynamics \revtext{as coupled ODEs, the proposed framework} enables the introduction of \revtext{physically meaningful} constraints on degradation that act as effective regularizers, leading to more generalizable and physically plausible degradation inference.
\revtext{The framework focuses on unsupervised degradation inference and can be applied to a single mechanical or infrastructure system given sufficiently long operational data, without requiring fleet-level data or supervision. When multiple run-to-failure trajectories are available, a natural extension is to leverage the inferred degradation states to perform fleet-level reliability analysis or supervised RUL prediction.}
Nevertheless, the current work \revtext{considers} degradation inference under a single failure mode, either as global stiffness loss in a bridge system or as a fault on a single component in an aero-engine system. A natural extension is to systems with multiple failure mechanisms, where different degradation processes may interact and influence each other's progression.
Future work could also explore several other promising directions. \revtext{First, the continuous-time formulation can be leveraged to naturally handle missing values and irregularly sampled data. Second, NCDE-based extensions could further investigate} log-CDE methods using log-signatures to improve memory over long horizons, as well as higher-order differential equations. \revtext{Lastly, for the learned differential operator in H-CDE,} explicit physics knowledge could be incorporated, and uncertainty \revtext{could be modeled via Neural Stochastic Ordinary Equations (Neural SDEs).}

\section{Data and Code Availability}
The script we used to generate this synthetic bridge dataset is available in the associated code repository to ensure reproducibility and facilitate further studies. Our code and data will be available upon acceptance under \url{https://github.com/EPFL-IMOS/hcde}.

\section*{Acknowledgment of AI Assistance in Manuscript Preparation}
During the preparation of this work, the authors used ChatGPT to assist with refining and correcting the text. After using this tool, the authors carefully reviewed and edited the content as needed and take full responsibility for the content of this publication.

\section*{Acknowledgments}
This work was supported by the Swiss National Science Foundation under Grant 200021\_200461. 


%

\appendix
\section{Details of Case Studies Setups}
\subsection{Bridge Simulation Parameters and Details
\label{app:bridge_params}}
This appendix provides the specific parameters and further details for the simulated degrading bridge case study discussed in Section~\ref{ssec:bridge-case}.

\textbf{Geometry and Discretizations.}
The bridge model has a total length $L = 10$\,m and is discretized into $N_e = 20$ finite elements (21 nodes). Boundary conditions are simply supported (pinned at node 0, roller at node 20).

\textbf{Material Properties.}
The initial material properties, representing Medium-Density Fiberboard (MDF), are: Young's modulus $E = 4.0 \times 10^9~\mathrm{Pa}$; moment of inertia $I = 0.0005~\mathrm{m^4}$; cross-sectional area $A = 0.06~\mathrm{m^2}$; density $\rho = 550~\mathrm{kg/m^3}$; and thermal expansion coefficient $\alpha = 5 \times 10^{-6}~\mathrm{K^{-1}}$.

\textbf{Damping Parameters.}
Rayleigh damping ($C = \alpha_{\text{damp}}\,M + \beta_{\text{damp}}\,K$) uses coefficients $\alpha_{\text{damp}}=0.1\,\mathrm{s}^{-1}$ (mass-proportional) and $\beta_{\text{damp}}=0.015\,\mathrm{s}$ (stiffness-proportional).

\textbf{Input Data Generation.}
\textit{Temperature:} Hourly data (0.1°C precision) from Zurich Fluntern weather station (via Visual Crossing) for 2023 and 2024.
\textit{Load:} Based on SBB passenger traffic statistics (Zurich main station, 2024 hourly percentages for short/long distance trains). Daily load factors are derived reflecting these patterns (with random daily variations, standard deviation 0.1). The distributed load $q(t)$ is obtained by applying these factors to a base scaling constant of 36 (resulting units nominally N/m).

\textbf{Damage Model Parameters.}
The damage accumulation follows $\Delta D = \beta_{\text{damage}} \, (1-D) ( (v_{\max} - U_{\mathrm{ref}}) / U_{\mathrm{ref}} )^p$ for $v_{max} > U_{ref}$. The parameters are: damage initiation threshold $U_{\mathrm{ref}} = 0.0125$\,m (L/800, max absolute vertical displacement); damage scaling parameter $\beta_{\text{damage}} = 3.2 \times 10^{-4}$ (applied per time step $\Delta t$); and damage exponent $p=2$.

\textbf{Thermal Load Calculation Details.}
Thermal loads contributing to $F(t)$ include axial forces from uniform temperature change $\Delta T = T_{ambient} - T_{\mathrm{ref}}$ (with $T_{\mathrm{ref}}=20^\circ\mathrm{C}$), and bending moments from an effective temperature gradient across the beam depth. This gradient is calculated via an exponential smoothing filter (update factor 0.1) applied to the recent ambient temperature history to represent thermal inertia, and is scaled by $\beta_{\text{thermal}}=1.0$.

\textbf{Numerical Integration Scheme.}
The Newmark--beta integration method (average acceleration scheme with $\beta = 0.25$, $\gamma = 0.5$) is employed. Integration uses a substep time interval $dt_\text{sub} = 1$\,min within each main input record interval $\Delta t = 10$\,min.

\textbf{Run-to-Failure Simulation Setup.}
Individual bridge instances start at different starting points, offset by 2 months to capture varying seasonal conditions. Each simulation proceeds until the bridge reaches end-of-life, defined as 30\% stiffness reduction ($D \approx 0.3$). Two primary input scenarios are modeled: Scenario A uses 2023 meteorological temperature data combined with a load pattern derived from long-distance traffic statistics, while Scenario B uses 2024 meteorological temperature data combined with a load pattern derived from short-distance traffic statistics (resulting in sharper daily peaks).

\section{Supervised Degradation Estimation}
\label{sec:eval_supervised_task} 
\revtext{{\textbf{Motivation}.}
While H-CDE is primarily designed for unsupervised degradation inference, most existing data-driven degradation models rely on supervised objectives such as ground-truth health state, degradation measurements, or run-to-failure trajectories. 
To complement the unsupervised evaluation and assess the architectural capability of H-CDE in capturing long-range dependencies and multiscale dynamics, we introduce a supervised degradation inference task on the Bridge case study.

\revtext{\textbf{H-CDE Configration.}} 
Although H-CDE is designed to model latent dynamics on two time resolutions, we adapt it here to infer latent states on a single time scale. Specifically, the input sequence is provided to the first CDE, and the second CDE takes the interpolated output path of the first CDE as its control signal to produce the degradation state.

\revtext{\textbf{Baseline Methods.}}
We compare H-CDE against representative time-series models selected for their complementary modeling properties.
CNN and GRU are included as commonly used baselines in supervised degradation estimation; TimeMixer represents recent multiscale sequence architectures; ODE-GRU captures long-range temporal dependencies in continuous time; and a standard NCDE serves as a reference model assess the contribution of the proposed hierarchical structure. 
The evaluated baselines are:
\begin{itemize}[leftmargin=1.5em, itemsep=2pt, topsep=4pt, partopsep=0pt, parsep=0pt]
    \item \textbf{CNN}: 1D-Convolutional Neural Network with an MLP regression head.
    \item \textbf{GRU}: Gated Recurrent Unit network with an MLP regression head.
    \item \textbf{TimeMixer}~\cite{wang_timemixer_2023}: MLP-based multiscale architecture adapted for regression by removing the decoder and adding a regression head, following~\cite{zhao_graph_2025}.
    \item \textbf{ODE-GRU}~\cite{rubanova_latent_2019}: A NODE extension with GRU updates at observation times and continuous ODE evolution between observations.
    \item \textbf{CDE}~\cite{kidger_neural_2020}: A single-layer vanilla Neural CDE model.
\end{itemize}

\revtext{\textbf{Training Setup.}}
The same training and test splits as in the unsupervised setting are used (see Sec.~\ref{ssec:bridge_split}). 
Degradation labels are provided as ground truth to all models during training.
All methods operate on identical input data consisting of sliding windows of length 100 with a stride of six.
Models are trained for up to 120 epochs with early stopping (patience of 10 epochs and a minimum of 20 epochs). The optimizer and learning rate follow the same configuration as in the unsupervised setting (see Sec.~\ref{sec:common_training_params}).

\textbf{Evaluation Metrics.}
Performance is evaluated using the Mean Absolute Error (MAE) and Root Mean Squared Error (RMSE) between predicted and true degradation values. In addition, we adopt the $\alpha$--$\lambda$ accuracy metric~\cite{saxena_metrics_2008} to assess prediction accuracy near the end of life, which is critical for downstream prognostics tasks. We set $\alpha=0.1$ and $\lambda=0.2$, focusing on the final 20\% of the time series and considering predictions accurate if the error remains within a 10\% tolerance.

\begin{table}[tbh]
\centering
\footnotesize
\caption{\revtext{Supervised degradation inference on bridge dataset (5 runs). Best results are in \textbf{bold}, second-best are \underline{underlined}.}}
\label{tab:bridge_supervised}
\begin{tabular}{lccl}
\toprule
Model & MAE $\downarrow$ & RMSE $\downarrow$  &$\alpha-\lambda \uparrow$ \\
\midrule
CNN & 0.019±0.004 & 0.025±0.007  &0.302±0.163 \\
GRU & 0.013±0.001 &  \underline{0.016±0.001}  &0.516±0.091 \\
TimeMixer & 0.013±0.002 & {0.017±0.002}  &{0.538±0.075} \\
ODE-GRU & 0.032±0.002 & 0.048±0.002  &0.094±0.039 \\
CDE & \underline{0.012±0.002} & {0.017±0.002}  &\underline{0.738±0.054} \\
\midrule
H-CDE & \textbf{0.008±0.001} & \textbf{0.010±0.001}  &\textbf{0.893±0.020} \\
\quad - w/o MC & 0.010±0.002 & 0.014±0.002  &0.837±0.053 \\
\quad - w/o PT & 0.015±0.005 & 0.019±0.006  &0.692±0.113 \\
\bottomrule
\end{tabular}
\end{table}

\textbf{Results and Discussion}.
As shown in Tab.~\ref{tab:bridge_supervised}, H-CDE achieves the best performance across all evaluation metrics. This demonstrates its strong ability to capture long-range temporal dependencies and multiscale dynamics, particularly for degradation prediction near end of life, as reflected by the $\alpha$--$\lambda$ accuracy.
The single-layer CDE achieves the second-best overall performance. However, H-CDE provides a substantial improvement over the single-layer CDE, reducing both MAE and RMSE by approximately 40\%, indicating that the hierarchical structure plays a critical role in improving degradation inference.
ODE-GRU performs poorly across all metrics, particularly on the $\alpha$--$\lambda$ score. This is largely due to its fixed NODE dynamics, which limit generalization to unseen operating conditions. As most test units lie in out-of-distribution regimes, these results highlight the advantage of NCDE-based models over NODE-based formulations under varying conditions.
Ablation results show that removing either the path transformation or the monotonic constraint degrades performance. The path transformation is particularly important, as it maps fast-timescale operational dynamics to a slower latent representation, facilitating the extraction of degradation-relevant drivers. The monotonic constraint also contributes to performance gains, especially during the early stages of degradation where changes are subtle, by providing a regularizer for the learned degradation dynamics.
}

\bibliographystyle{elsarticle-num} 
\bibliography{bibtex/bib/references}
%


\end{document}